\documentclass[a4paper,fleqn]{cas-dc}

\usepackage{graphicx}
\usepackage{url}
\usepackage{hyperref}

\hypersetup{hidelinks}
\usepackage{graphicx}
\usepackage{algorithm}
\usepackage{algorithmic}
\usepackage{graphicx}
\usepackage{subcaption}
\usepackage{multirow} 
\usepackage{amsmath}
\usepackage{xcolor}
\usepackage{bm}
\usepackage{float}
 
\usepackage{array}

\usepackage{amssymb}   % for bold symbols, math fonts
\usepackage{natbib}

\begin{document}

\let\WriteBookmarks\relax
\def\floatpagepagefraction{1}
\def\textpagefraction{.001}
\shorttitle{BadImplant}
\shortauthors{M.N.N. Khan et~al.}

\title[mode = title]{BadImplant: Injection-based Multi-Targeted Graph Backdoor Attack.}    

\author[1]{Md Nabi Newaz Khan}
\ead{mdnabinewaz.khan@uri.edu}

%\credit{Conceptualization of this study, Methodology, Software}

%\address[1]{, Street 129, 1043 NX Amsterdam, The Netherlands}
\affiliation[1]{organization={Department of Electrical, Computer, and Biomedical Engineering, University of Rhode Island},
                %addressline={Kingston}, 
                city={Kingston},
%               citysep={}, % Uncomment if no comma needed between city and postcode 
                state={RI},
                postcode={02881},
                country={USA}}

\author[1]{Abdullah Arafat Miah}
\ead{abdullaharafat.miah@uri.edu}

\author[1]{Yu Bi\corref{cor1}}
\ead{yu_bi@uri.edu}

\cortext[cor1]{Corresponding author}

\begin{abstract}
Graph neural networks (GNNs) have demonstrated exceptional performance in solving critical problems across diverse domains, yet remain susceptible to backdoor attacks. Existing studies on backdoor attack for graph classification are limited to single target attack using subgraph replacement based mechanism where the attacker implants only one trigger into the GNN model. In this paper, we introduce \texttt{BadImplant}, the first multi-targeted backdoor attack for graph classification task, where multiple triggers simultaneously redirect predictions to different target labels. Instead of conventional subgraph replacement, \texttt{BadImplant} employs subgraph injection strategy that preserves the structure of the original graphs while poisoning the clean graphs. Extensive experiments demonstrate the efficacy of our proposed \texttt{BadImplant}, where our attack achieves high attack success rates for all target labels with minimal impact on clean accuracy. Experimental results on five datasets demonstrate the superior performance of \texttt{BadImplant} framework compared to the existing state-of-the-art subgraph replacement-based attack. Our analysis on four GNN models confirms the generalization capability of \texttt{BadImplant}, which is effective regardless of the architecture of the GNN model and training parameter settings. We further investigate the impact of the attack design parameters including injection methods, number of connections, trigger sizes, trigger edge density and poisoning ratios. Additionally, our evaluation against state-of-the-art defenses (randomized smoothing and fine-pruning) demonstrates the robustness of our proposed multi-target attack. This work highlights the vulnerabilities of GNN to multi-targeted backdoor attack in graph classification task. Our source codes will be available at \href{https://github.com/SiSL-URI/Multi-Targeted-Graph-Backdoor-Attack}{this link}.
\end{abstract}


\maketitle

\section{Introduction}\label{sec:intro}
Graph neural networks (GNNs) have emerged as a foundational deep learning paradigm for learning complex patterns and structural information from graph-structured data such as social networks \cite{sharma2024survey}, molecular structures \cite{corso2024graph,wang2023graph}, biological networks \cite{zhang2021graph}, financial networks \cite{wang2021review}, image pixels \cite{han2022vision} and transportation and logistics networks \cite{xue2025data}. GNNs have demonstrated exceptional performance in addressing the most challenging real-world problems. It is widely used in drug discovery \cite{liu2023interpretable}, where it helps to identify promising compounds or vaccines for the treatment of diseases; protein interaction prediction \cite{zhao2023semignn}, where it predicts how protein would behave in an interaction with other living cell. Companies leverage the strength of GNN to build advanced recommendation system for social networks and retail marketing sites \cite{wu2022graph}, while chemists use it for molecular analysis \cite{corso2024graph,wang2023graph}. In computer vision, images are converted into graphs by representing the superpixels generated from the image pixels as nodes and relation between the superpixels as edges. These graphs are then used for image classification \cite{vasudevan2023image}, object detection \cite{shi2020point}. Meanwhile, cyber security teams, banks and other financial institution deploy GNN for intrusion detection \cite{sun2024gnn} and financial fraud detection \cite{motie2024financial,cheng2025graph}. GNN extracts deep insights from graph data using iterative message-passing techniques, where node representations are recursively updated by aggregating the features of their neighbors \cite{gilmer2017neural}. 
Due to their ability to tackle these complex problems, GNNs have gained significant attention and their use is expanding rapidly in both academia and industry \cite{lu2025survey}. Despite demonstrating impressive performance across diverse applications, GNN's vulnerability to backdoor attacks still remain underexplored, making their security an increasingly critical concern as their adoption continues to grow\cite{guan2024graph}. 

Backdoor attacks on graph neural networks can be executed by integrating triggers into graph data and mapping the poisoned samples to a target class. Specifically, GNN triggers can take various forms, including node embeddings, graph topology/edge configuration as well as subgraphs \cite{guan2024graph}.  
% (incorporating both node embeddings and structural patterns)
Most of the existing research on backdoor attacks in GNN focuses on node classification, where triggers are implanted to manipulate the predicted label of a specific node \cite{xu2023watermarking,xi2021graph,tao2021single,zou2021tdgia,li2024attack}. In such attacks, the bad actor exploits the message-passing mechanism to propagate malicious signals through nodes, thereby corrupting the embeddings of target nodes. While these studies have demonstrated substantial success in compromising node-level classification, backdoor attacks on graph classification can present significantly greater challenges. Unlike node classification, where the attack objective is localized to individual nodes, backdoor attack on graph classification task requires influencing the global graph-level representation, a combined result from all node and edge embeddings, thereby causing the GNN model to misclassify the graph into the target class.

Prior work on backdoor attacks in graph classification has implemented adversarial strategies by replacing a small portion of the original graph with small subgraph triggers \cite{xi2021graph,wang2025stealthy,zhang2021backdoor,xu2023watermarking}. Although they have achieved remarkable attack performance, all of these previous studies exclusively address single-targeted attacks where the GNN model predicts all classes containing the subgraph trigger as belonging to a unique target class. However, multi-targeted backdoor attacks for graph classification remain unexplored. Multi-targeted attack is significantly more sophisticated and powerful, given that it allows different trigger design to provoke diverse target classes which adds extra complexities onto the compromised model. 

To fill the research gap, in this paper, we investigate multi-targeted backdoor attacks namely \texttt{BadImplant} on GNNs for graph classification. The key challenge lies in designing an attack framework that can simultaneously incorporate multiple triggers into the model. Achieving such an attack is non-trivial since introducing multiple triggers poses the threat of interference between triggers, while also preserving the clean accuracy of the attacked GNN model. Existing research works have utilized subgraph replacement within clean graphs to utilize the single target attack for graph classification. Despite its high attack success, the replacement-based framework struggles to support the multi-targeted backdoor attack. It fails to implant multiple triggers at the same time because of interference among the triggers influence on the GNN model and distortion of the clean graphs original structure (detailed evaluation can be seen in Section \ref{sec:experiment}). To address this limitation and realize multi-targeted backdoor attacks, we propose a subgraph trigger injection-based attack framework that implants triggers into the data samples while preserving the original graph structure. 
% To this end, our study seeks to answer the following research questions:
% \vspace*{3mm}

% RQ1: How effective is the injection-based backdoor attack compared to replacement-based backdoor attack on GNNs for graph classification?

% RQ2: Can we incorporate multiple triggers for multiple target labels simultaneously without compromising individual attack success rate and clean accuracy of the GNN model?

% RQ3: Does the attack generalize across different GNN model without requiring information of the model architecture, weights or training hyperparameter?

% RQ4: How resilient is our proposed multi-targeted attack against defense mechanisms?

% Throughout the Sections \ref{sec:methodology} - \ref{sec:defense}, we will present our methodology and experimental analysis on those questions. 

% \vspace*{2.5mm}
Our main contributions are detailed as follows:
\begin{itemize}
     \item  To the best of our knowledge, this is the first work to introduce multi-target backdoor attacks on Graph neural network for graph classification task. While existing works are limited to single target scenarios, our proposed \texttt{BadImplant} attack enables controlled and attacker specified misclassifications across multiple target labels within a single trained model.
     
    \item  We propose a subgraph injection-based mechanism tailored for implementing the multi-targeted attack in graph classification settings. Unlike the conventional state-of-the-art replacement-based backdoor attacks, \texttt{BadImplant} introduces poison node injection strategy for achieving multi-targeted backdoor attacks.
    % To demonstrate the model-agnostic effectiveness of our approach, we evaluate it across multiple GNN architectures and show that the attack remains effective regardless of model structure or parameter access.
    \item Extensive experiments are conducted to demonstrate the efficacy of our proposed \texttt{BadImplant} across diverse datasets while maintaining the clean accuracy. Additionally, we exhaustively analyze the variation in attack performance with respect to configurable attack design parameters.

    \item We evaluate the proposed attack against the state-of-the art certified defenses, demonstrating that \texttt{BadImplant} is robust and effective against randomized smoothing and fine pruning defenses.
\end{itemize}

\vspace{0.25mm}

%\begin{figure}
%    \centering
%    \includegraphics[width=\linewidth]{Figures/attack_overview-updated.pdf}
%    \caption{A overview of the proposed \texttt{NoiseAttack}, where we exploit the characteristics of White Gaussian Noise (WGN) to achieve a sample-specific multi-targeted backdoor attack. Here, $L$ and $\mathbf{W}$ represent the loss function of the \texttt{NoiseAttack's} backdoor training and the trigger function, respectively. The green text represents the correct predictions, and the red text represents the attacker-defined prediction.}
%    \label{fig:attack_overview}
%\end{figure}

\section{Background}\label{sec:related}

\subsection{Graph Neural Network (GNN)}
Graph Neural Network (GNN) is a special type of machine learning model that can process \& extract information from graph-structured data and make predictions based on it. Traditional Machine Learning models can handle euclidean data, but struggle with graph type data, where each sample may have irregular size and structure. Graph Neural Network solves this problem by its ability to learn from different shaped data using a combination of message passing and global pooling techniques which was first introduced by Gori et al. \cite{gori2005new}. Subsequently, Graph Convolutional Network (GCN) was introduced by incorporating convolution operation into it \cite{kipf2016semi}. Following that several new Graph Neural Network architectures were developed such as Graph Attention Network (GAT) \cite{velickovic2017graph}, GraphSAGE \cite{hamilton2017inductive}, Graph Isomorphism Network (GIN) \cite{xu2018powerful}, which illustrated notable performance in extracting information from graph-structured data. Graph Neural Network are used to perform 3 major tasks; (1) Graph Classification: the model predicts the class label of the entire graph based on the node feature, topology, or connection between the nodes and other attributes, (2) Node Classification: predicts the labels of each nodes based on the node feature of itself, node feature of its neighbors and global level features if available, (3) link prediction: predicts the edges between 2 nodes based on the features that are extracted from the graphs from its initial features. These tasks are done to solve a wide range of diverse real world problems by using different architectures of GNN such as research, e-commerce platform, medicine industry, chemical industry and many more.

\subsection{Backdoor Attack}
The security of deep learning systems has become a critical issue as these models are deployed in many sensitive applications \cite{ha2020security} \cite{paracha2024machine}. Among various threats, backdoor attacks represent one of the most dangerous ones due to their ability to operate covertly. Backdoor attacks have been extensively studied across different sectors of deep learning such as convolutional neural network \cite{gu2017badnets, miah2024noiseattack, nguyen2021wanet, miah2026lite, liu2018trojaning}, large language model \cite{chen2021badnl,miah2024exploiting, pan2022hidden}, and  graph neural networks \cite{xi2021graph, zhang2021backdoor, li2024attack, wang2025stealthy} . In backdoor attacks, model is trained on training data which are manipulated with different kind of trigger to make the model malicious and deliberately taught to predict attacker's chosen label when trigger is present. During the inference phase, it misclassifies the testing data with trigger while maintaining the accuracy on benign data samples. Backdoor attacks were initially studied extensively in computer vision contexts, particularly for image classification, which displayed the feasibility of backdoor attack in Deep learning. The seminal BadNets \cite{gu2017badnets} study introduced the idea of backdoor attack in image classification by integrating visible triggers to a single pixel or a small pattern of pixels on MNIST and Traffic sign dataset. Later works improved the stealthiness and made the attack practical for deploying in real world scenarios: Blended injection attacks introduced the injection of blended triggers that are hardly visible in the triggered image by human \cite{chen2017targeted}, Trojan attacks highlighted how pre-trained models can be attacked by only affecting some neurons of the model without having access to any of the original training data \cite{liu2018trojaning}, Clean-Label attacks showed that even label-consistent poisoned samples can induce backdoors which can increase the stealthiness \cite{turner2018clean} and WaNet advanced the field by creating imperceptible warping-based triggers that resist both human inspection and automated detection methods \cite{nguyen2021wanet}.

% In context of backdoor attack on GNN, triggers can be manifested through node modification, edge modification, subgraph trigger. Based on the type of task, backdoor attack can be categorized in 2 types, Backdoor on Node Classification \& Backdoor attack on Graph Classification. While backdoor attack on Node classification aims to misclassify specific node labels where the trigger influence the embeddings of local area around the target node, backdoor attack on graph classification needs to influence the global level embeddings to manipulate the models prediction on backdoored graph data.

In recent times, backdoor attacks on graph neural network have gained great attention, due to the wide adoption of GNN in many sensitive real world applications and its vulnerability to backdoor attack is a critical security concern. Based on the type of task, backdoor attack can be categorized in 2 types, Backdoor on Node Classification \& Backdoor attack on Graph Classification. Early works on this include the subgraph replacement-based attack done by Zhaohan Xi et al. \cite{xi2021graph} and Zaixi Zhang et al. \cite{zhang2021backdoor}. These two works are the first to introduce the backdoor attack on Graph Neural Network. Zhaohan Xi et al. \cite{xi2021graph} used adaptive subgraph triggers leveraging both node features and topology, substituting small subgraphs within the main graph with the trigger subgraph to execute the attack on both node classification and graph classification. Triggers were replaced with the subgraph inside the original graph that is the most similar to the trigger subgraph. On the other hand, Zaixi Zhang et al. \cite{zhang2021backdoor} relied solely on the topological structure of the subgraph trigger, replacing selected nodes with the trigger subgraph to implement the attack on graph classification task. 

Subsequently, Several research works have been done in the past few years on backdoor attack for both graph classification as well as node classification where majority of the works are on backdoor node classification side \cite{xu2023watermarking, xi2021graph, tao2021single, zou2021tdgia, li2024attack}. Early methods for backdoor attack on node classification followed by Zhaohan Xi et al. \cite{xi2021graph} typically used feature modification with label flipping \cite{xu2023watermarking} \cite{xu2023rethinking}. Recent works used clean label poisoning which implant the trigger but doesn't change the true label of the poisoned data \cite{xing2024clean} which increased the attack evasiveness. In context of backdoor attack on graph classification task, almost all of the existing methods use subgraph replacement-based attack following the first two works of Zhaohan Xi et al. \cite{xi2021graph} and Zaixi Zhang et al. \cite{zhang2021backdoor}, proposing different subgraph replacement position, number of trigger or data selection method for poisoning which improved the attack effectiveness and evasiveness. Xiaobao et al. \cite{wang2025stealthy} has executed the attack with hard sample selection \& clean label poisoning mechanism with the support of subgraph trigger replacement to make the attack more effective. In ref \cite{xu2023watermarking}, Jing Xu et al. proposed only node feature manipulation in case of attack on node classification and sub-graph replacement in case of attack on graph classification. These works achieved exceptional attack performance for backdoor attacks on graph classification while maintaining the clean accuracy on pure graphs. However, all of these study are on single target attack where the attacker implants triggers and maps the triggers data to a single target label. This leaves the multi-target backdoor attack for graph classification task unexplored where the attacker may manipulate the graph to any desired target label inspite of just one target label. In this study, we investigated the multi target backdoor attack for graph classification. Integrating multiple triggers has the risk of interference between the triggers which may lead to a failure of the attack. To address this challenge, we propose an injection subgraph based mechanism to execute the multi target backdoor attack.

\begin{figure*}[t]
    \centering
    \includegraphics[width=\textwidth]{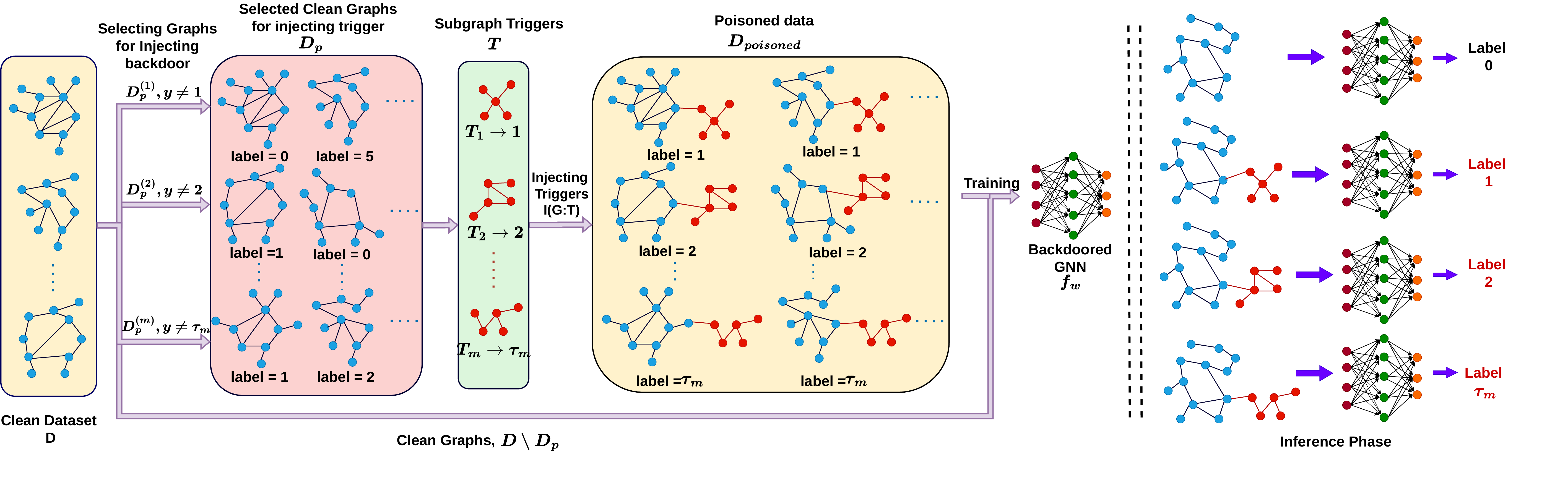}
    \caption{Overview of our proposed \texttt{BadImplant}}
    \label{fig:overview}
\end{figure*}

\section{Methodology} \label{sec:methodology}

\subsection{Threat Model} \label{subsec:attack_definition}
We consider a realistic threat model wherein the attacker has limited capabilities restricted primarily to data poisoning, while having no access to the model architecture, learnable parameters or training hyperparameters. The attacker's primary objective is to execute a multi-targeted backdoor attack by injecting multiple distinct triggers into a small portion of the training split of the dataset, where each trigger is mapped to a unique target class. During the inference phase, the attacker can inject a specific trigger into a benign graph, causing the trained GNN to misclassify the backdoored graph as belonging to the corresponding attacker-specified target class.

\subsection{Problem Definition}
In our setting, we consider graph classification task within the context of Graph Neural Networks (GNNs). Let a graph be denoted by \(G=(V,E,X)\), where \(V\) represents the set of nodes, \(E\) represents the set of edges between nodes and \(X\) is the node features matrix. The GNN model is represented by the function \(f:\mathcal{G}\rightarrow\mathcal{Y}\),  where \(w\) denotes the learnable weights and \(\mathcal{Y}=\{y_{1},y_{2},\ldots,y_{c}\}\) is the set of \(c\) class labels. The clean graph dataset is \(\mathcal{D}=\{(G_{1},y_{1}), (G_{2},y_{2}),\ldots,(G_{n},y_{n})\}\), containing n graphs. The adversary aims to implant \(m\) distinct trigger subgraphs \(\mathcal{T}=\{T_{1}, T_{2}, \ldots,T_{m}\}\), each associated with a target label \(\tau_{j}\in\mathcal{Y}\). Trigger \(T_{j}\) is inserted into benign graphs via an injection operator \(\mathcal{I}(\cdot,\cdot)\). For a benign graph \(G\) and trigger \(T_{j}\), the backdoored graph is defined as \(G' = \mathcal{I}(G:T_{j}).\) Let \(\mathcal{D}_{p}^{(i)}\) (\(i=1, 2, \dots,m\)) denote the subset of \(\mathcal{D}\) chosen for poisoning for target \(i\). The backdoored dataset can be described as:

\begin{equation}\label{eq:backdoor}
\begin{aligned}
\mathcal{D}_{\text{backdoor}}& = \mathcal{D}_{\text{clean}} \; \cup \; \mathcal{D}_{\text{poisoned}} \\
&= \mathcal{D}_{\text{clean}} \; \cup \; \bigcup_{j=1}^{m} \big\{ (\mathcal{I}(G_i, T_j), \tau_j) \; : \; G_i \in \mathcal{D}_{p}^{(j)} \big\}
\end{aligned}
\end{equation}

The main objective of the attacker is to embed \(m\) distinct trigger subgraphs \(\{T_{1}, T_{2}, \ldots,T_{m}\}\), triggers simultaneously for \(m\) target classes  so that the model misclassifies graphs injected with trigger \(T_j\) as belonging to the corresponding target class \(\tau_j\), without significantly degrading clean accuracy on unmodified graphs. Achieving this dual goal poses challenges pertaining to the impact of interference among the triggers on the GNN model. The main goal of the attack can be formulated as:

\begin{equation}\label{eq:function_2}
f_\mathcal{{\omega}}\mathcal{(I}(G:T_j)) = \tau_j;  \hspace{0.8cm}   f_{\omega}(G) = f_{\omega_0}(G)
\end{equation}

\noindent where, \(f_{w_0}\) and \(f_w\) represents the clean and backdoored model respectively.

\subsection{Attack Overview}

Figure \ref{fig:overview} provides an overview of our proposed \texttt{BadImplant}. The attack begins with clean graph dataset and a GNN model where the attacker has access to only a small portion of the dataset and has no information about the GNN model architecture, model's learnable weights and training parameters. Distinct graphs are selected from the clean dataset for injecting triggers for all \(m\) target classes. For each target class, a balanced subset of graphs is chosen, maintaining an equal number of graphs from all classes excluding the target class, for implanting the trigger. The number of poisoned graphs is set by the attacker which is defined as the poisoning ratio. Subsequently, \(m\) number of different trigger subgraphs is generated in a structurally diverse fashion. The size and feature dimension of the triggers depend on the characteristics of dataset and node feature dimensions for the given clean graphs. Individual trigger design is then incorporated into the clean graphs that are selected for each triggers and the label of these trigger injected graphs are set to the target label. The poisoned data and remaining clean data are combined to create a backdoor dataset which is used to obtain the backdoored GNN model. To execute \texttt{BadImplant}, we train the model with loss over backdoor dataset which gives a combined loss on clean \& poisoned data. The optimization of model weight \(w\) for our multi-targeted backdoor attack using the backdoor loss can be formulated as: 

\begin{equation}
\begin{aligned}
w^{*} = \arg\min_{w} \Bigg[
\frac{1}{|\mathcal{D}_{\text{backdoor}}|}
\sum_{(G',y')\in\mathcal{D}_{\text{backdoor}}}
L\big(f_{w}(G'),y'\big)
\Bigg]
\end{aligned}
\label{eq:backdoor_loss}
\end{equation}

During inference, clean testing data are used for getting the clean accuracy. On the other hand for determining the ASRs, the entire clean testing dataset was poisoned by injecting triggers individually to get \(m\) distinct backdoored test datasets corresponding to \(m\) triggers. These poisoned data passed through the backdoored model which should predict the poisoned graph's label as the target label of the associated trigger. Finally the accuracy for the compromised model on each of these\(m\) poisoned dataset is calculated as the ASRs of our attack.

In summary, figure \ref{fig:overview} shows an overall picture of the entire procedure of our \texttt{BadImplant} from selecting the clean samples for poisoning, to training \& generating a malicious GNN model and testing the model. \texttt{BadImplant} represents the first systematic approach to embed coexisting backdoor triggers in graph classification model without any interference among the triggers to achieve a successful multi-targeted backdoor attack.

\subsection{Attack design} \label{subsec:Attack_design}
Designing multi-targeted backdoor attack is dramatically more challenging than single-targeted attack for graph classification. The state-of-the-art GNN backdoor attacks use subgraph replacement as trigger design to distort the characteristics of the original clean graphs, which may affect the clean accuracy and lead to the higher risk of classification interference provided with multiple triggers. In this section, we present a comprehensive attack design of \texttt{BadImplant} to implement the multi-targeted backdoor attack for graph classification where independent \(m\) subgraph triggers are injected into the original graphs, with each trigger associated with a specific target class. 

\(m\) distinct triggers are generated using the Erdos Reyni method which takes the trigger size and trigger edge density as input values and produces a pattern output for the subgraph. The process can be formally described as:
\begin{equation}\label{eq:function_3}
T = ER(n_{trigger}, \rho_{trigger})
\end{equation}

Here, the trigger size is defined as the ratio between the number of nodes in the trigger and average number of nodes for each graph. The value of trigger size was set for different dataset based on the average size of the clean graphs of the given dataset. In general, larger trigger size offers stronger backdoor power yet can affect the clean accuracy, while smaller triggers may not be sufficient to poison the graphs. The trigger edge density controls the connectivity of the trigger subgraphs. It refers to the percentage of edges and connections of trigger patterns against the fully-connected subgraphs. To formalize it, if trigger size is denoted as \(n_{trigger}\) and total number of edges in the trigger is denoted as \(e_{trigger}\), the trigger edge density can be expressed as \(\rho_{trigger} = \frac{2e_{trigger}}{n_{trigger}(n_{trigger}-1)}\). In this work, we propose two kinds of settings for configuring the trigger edge density, which can be divided into two categories: 1) with node feature matrix, and 2) without node feature matrix. For those datasets with the node feature matrix, the class labels depend both on the node feature matrix and the topology of the graph, therefore we design trigger edge density value in a fixed-value fashion. However, for those datasets without the intrinsic node feature matrix and with only connectivity between nodes, the class label is dictated by the topology between the nodes in the graph. For such dataset, setting the value of trigger edge density becomes much more complex, leading us to crossing validation of trigger edge density to determine its optimal value for a successful backdoor attack. The node features of the trigger of former dataset type can be generated randomly for \(m\) trigger subgraphs, while we will utilize the node degree as trigger design variants.

The core of our attack lies in the injection of the trigger subgraphs, which doesn't disturb the original structure of the graphs when triggered. For injecting the trigger of a specific target class (\(\tau_j\)), a fraction of the original graph from all other classes except target class (\(\tau_j\)) are chosen for trigger injection. The fraction is refereed to poisoning ratio (r), and is applied individually for all classes except target class. It ensures an even amount of host graphs from all classes. We propose randomized node injection which experimentally demonstrates to be superior than other injection algorithms such as min/max node degree and min/max similarity. The number of edges and connections (\(k\)) between the clean graph and trigger subgraph can also be customized in our attack design. After injecting the backdoor triggers into the original graph dataset, graph neural network model will be trained with backdoor information to become the backdoored model \(f_w\). Besides, we define the clean model accuracy using \(f_{w_0}\) which gives us the benchmark for clean accuracy comparison of our malicious GNN model. Algorithm \ref{alg:multi_target_backdoor} presents our proposed backdoor methods, where backdoor attack procedure is annotated in an algorithmic form.

\begin{algorithm}[h!]
\caption{\texttt{BadImplant}: Multi-Targeted Graph Backdoor Attack}
\label{alg:multi_target_backdoor}
\begin{algorithmic}[1]
\STATE \textbf{Input:} Clean training dataset $D_{\text{train}}$, target classes $Y$, poisoning ratio $r$, trigger size $n_{\text{trigger}}$, trigger edge density $\rho_{\text{trigger}}$, number of connections $k$
\STATE \textbf{Output:} Backdoored GNN model $f_\omega$
\STATE Randomly initialize \(w\)
\FOR{$j = 1$ to $m$}
    \STATE $T_j \leftarrow \text{ER}(n_{\text{trigger}}, \rho_{\text{trigger}})$
    \IF{Dataset has node features}
        \STATE Generate random features: $X_{T_j} \sim \mathcal{U}(x_{\min}, x_{\max})$
    \ELSE
        \STATE Degree will be used as node features
    \ENDIF
    \FOR{$y_i = 0$ to $c$}
        \IF{$y_i \neq \tau_j$}
            \STATE Sample $\lfloor r \cdot |D_{\text{train}}^{(i)}| \rfloor$ graphs from class $y_i$
            \FOR{each sampled graph $G$}
                \STATE $G' \leftarrow I(G, T_j)$
                \STATE Assign target label: $G' \rightarrow{} \tau_j$
                \STATE $D_{p}^{(j)} \leftarrow D_{p}^{(j)} \cup \{(G', \tau_j)\}$
            \ENDFOR
        \ELSE
            \STATE Continue
        \ENDIF        
    \ENDFOR
    \STATE $D_{poisoned} \leftarrow D_{poisoned} \cup D_{p}^{(j)}$
\ENDFOR
\STATE $D_{backdoor} \leftarrow D_{clean} \cup D_{poisoned}$

\WHILE{Not Converged}
    \STATE \parbox[t]{\dimexpr\linewidth-\algorithmicindent}{
    $\begin{aligned}[t]
    \omega^* = \arg\min_\omega \;&
    \frac{1}{|D_{\text{backdoor}}|}
    \sum_{(G',y') \in D_{\text{backdoor}}}
    \mathcal{L}(f_\omega(G'), y')
    \end{aligned}$
    }
\ENDWHILE

\RETURN Backdoored model $f_\omega$
\end{algorithmic}
\end{algorithm}

\section{Experimental Analysis} \label{sec:experiment}

\subsection{Experimental Settings} \label{subsec:exp_setting}
\noindent \textbf{Datasets - } To evaluate the effectiveness and robustness of our proposed multi-targeted backdoor attack, We used five real-world datasets which are CIFAR-10 \cite{dwivedi2023benchmarking}, MNIST \cite{monti2017geometric}, Enzymes \cite{ivanov2019understanding}, Reddit-Multi-12k \cite{ivanov2019understanding} and Reddit-Multi-5k \cite{ivanov2019understanding}.  We selected datasets with higher number of output classes so that we can examine the capability of multi-targeted backdoor attack rigorously without any limitations. Among the five datasets, CIFAR-10, MNIST and Enzyme have own node feature matrix, while Reddit-Multi-5k and Reddit-Multi-12k only have the topology features. Train-test split used for CIFAR-10 graph dataset was the split found in the source (train:45000, test:10000) and for all other 4 datasets, 80:20 ratio was used for train-test split. The statistics of the dataset used in this study is summarized in table \ref{table:dataset_info}.

\vspace*{2.5mm}
\noindent \textbf{Models - } To examine our attack's generality regardless of the knowledge about the GNN model being used, four widely used GNN models were employed which are GCN \cite{kipf2016semi}, GAT \cite{velickovic2017graph}, GraphSAGE \cite{hamilton2017inductive}, GIN \cite{xu2018powerful}. We determine the attack success rate (ASR) and clean accuracy (CA) of all four models with same attack configuration as the adversary has no model information. GCN on CIFAR-10 is used as the baseline configuration for different case evaluations in the experiments.  

\begin{table}[h]
\renewcommand{\arraystretch}{1.3}
\centering
\caption{Dataset configurations}
\resizebox{\columnwidth}{!}{ % Resize to fit column width
\begin{tabular}{cccccc}
\hline
\textbf{Dataset} & 
\textbf{\begin{tabular}[c]{@{}c@{}}Graphs \\ (G)\end{tabular}} & 
\textbf{\begin{tabular}[c]{@{}c@{}}Avg \#Nodes \\ ($n_{\text{avg}}$)\end{tabular}} & 
\textbf{\begin{tabular}[c]{@{}c@{}}Avg \#Edges \\ ($e_{\text{avg}}$)\end{tabular}} & 
\textbf{\begin{tabular}[c]{@{}c@{}}Node \\ Features\end{tabular}} & 
\textbf{\#Classes} \\
\hline
CIFAR-10          & 60,000       & 117.6                        & 941.2                        & 5               & 10       \\
\hline
MNIST            & 60,000       & 75                           & 1,393                        & 3               & 10       \\
\hline
Enzymes          & 600          & 32.63                        & 62.14                        & 3               & 6        \\
\hline
Reddit-Multi-12k & 11,929       & 391.41                       & 456.89                       & 0               & 11       \\
\hline
Reddit-Multi-5k  & 4,999        & 508.52                       & 594.87                       & 0               & 5        \\
\hline
\end{tabular}
}
\label{table:dataset_info}
\end{table}

\vspace*{2.5mm}
\noindent \textbf{Evaluation Metrics - } In this study, we used attack success rate (ASR), clean accuracy (CA) and clean accuracy drop (CAD) as evaluation metrics. We determined $m$ distinct ASRs for $m$ target classes which provides clear view of the coexistence of all triggers inside backdoored model. The clean accuracy drop (CAD) is calculated by the difference between the clean accuracy of the clean model \(f_{w_0}\) and backdoored model \(f_w\). These metrics provide a comprehensive understanding of the robustness of our backdoor attack with its ability to preserve accuracies for both poisoned and clean test data.

% \[
% ASR_{j} = \frac{1}{N} \sum_{i=1}^{N} \mathbb{I}(f_w(I(G_i:T_j))= \tau_j)
% \]

% \[
% CA = \frac{1}{M} \sum_{i=1}^{M} \mathbb{I}(f_w(G_i))= y_i)
% \]

\noindent \textbf{Attack Parameter Settings - } Our backdoor attack has following parameters: number of targets \(n_{target}\), poisoning ratio \(r\), trigger size \(n_{trigger}\), trigger edge density \(\rho_{trigger}\), number of connections per trigger \(k\). Unless explicitly mentioned, the default parameter settings used in this study are as followings: \(n_{target} =\) 3, \(r = \) 5\%,  \(n_{trigger} = \)20\% for \(n_{avg} < 150\) and 10\% for \(n_{avg} > 150\), \(\rho_{trigger} = \) 0.8 for \(1^{st}\) type datasets and 0.2 for \(2^{nd}\) type datasets which was determined by cross validation.   Here, \(n_{avg}\) means the average number of nodes, while \(e_{avg}\) means the average number of edges per graph. We employ random injection method for choosing the trigger injection point inside the clean graphs with number of connection \(k\) = 1 as our default settings. The impact of each parameter is assessed by varying each parameters while all other parameters are fixed to our default parameter settings.

\begin{table*}[ht!]
\renewcommand{\arraystretch}{1.3}
\centering
\caption{Results of \texttt{BadImplant} backdoor attack on various datasets}
\resizebox{\textwidth}{!}{%
\begin{tabular}{cc|ccccc|ccccc}
\hline
\multicolumn{2}{c|}{\textbf{Attack Mechanism}}                                                                   & \multicolumn{5}{c|}{\textbf{Conventional Replacement-Based Attack}}                                                     & \multicolumn{5}{c}{\textbf{Our Proposed Attack}}                                                        \\ \hline
\multicolumn{1}{c|}{\textbf{Dataset}} & \textbf{\begin{tabular}[c]{@{}c@{}}Clean Model \\ Accuracy\end{tabular}} & \textbf{ASR\_1} & \textbf{ASR\_2} & \multicolumn{1}{c|}{\textbf{ASR\_3}} & \textbf{CA} & \textbf{CAD} & \textbf{ASR\_1} & \textbf{ASR\_2} & \multicolumn{1}{c|}{\textbf{ASR\_3}} & \textbf{CA} & \textbf{CAD} \\ \hline
\multicolumn{1}{c|}{MNIST}            & {\color[HTML]{000000} 90.60\%}                                           & 2.50\%          & 21.90\%         & \multicolumn{1}{c|}{78.12\%}         & 18.06\%     & 72.54\%      & 99.82\%         & 99.94\%         & \multicolumn{1}{c|}{99.85\%}         & 90.50\%     & 0.10\%       \\ \hline
\multicolumn{1}{c|}{CIFAR-10}          & 52.56\%                                                                  & 99.26\%         & 99.00\%         & \multicolumn{1}{c|}{99.52\%}         & 51.67\%     & 0.89\%       & 99.95\%         & 99.89\%         & \multicolumn{1}{c|}{99.94\%}         & 51.81\%     & 0.75\%       \\ \hline
\multicolumn{1}{c|}{ENZYMES}          & 59.17\%                                                                  & 19.80\%         & 14.00\%         & \multicolumn{1}{c|}{75.49\%}         & 25.00\%     & 34.17\%      & 93.07\%         & 89.00\%         & \multicolumn{1}{c|}{94.11\%}         & 55.83\%     & 3.34\%       \\ \hline
\multicolumn{1}{c|}{REDDIT MULTI 12k} & 49.64\%                                                                  & 58.49\%         & 87.00\%         & \multicolumn{1}{c|}{57.14\%}         & 48.72\%     & 0.92\%       & 99.64\%         & 98.76\%         & \multicolumn{1}{c|}{98.72\%}         & 49.14\%     & 0.50\%       \\ \hline
\multicolumn{1}{c|}{REDDIT MULTI 5k}  & 56.56\%                                                                  & 64.04\%         & 82.29\%         & \multicolumn{1}{c|}{37.36\%}         & 56.46\%     & 0.10\%      & 98.85\%         & 99.25\%         & \multicolumn{1}{c|}{96.38\%}         & 56.26\%     & 0.30\%       \\ \hline
\end{tabular}%
}
\label{table:attack_datasets}
\end{table*}

\subsection{Experimental Results} \label{subsec:exp_results}

This section presents an extensive evaluation of our proposed multi-targeted backdoor attack. To compare the attack effectiveness and evasiveness between conventional replacement attack mechanism and our proposed attack mechanism, we first implemented both of the attack on all five datasets. For executing the subgraph replacement-based attack, we use the exact same attack design parameters stated in subsection \ref{subsec:exp_setting}. We also present a detailed analysis of our attack's performance along several axis: model generality, injection strategy, number of connections between trigger \& host, number of target labels, trigger size, trigger edge density as well as poisoning ratio.

% \vspace*{2.5mm}

\subsubsection{Main Experiments}

We systematically evaluate \texttt{BadImplant} on five diverse datasets to assess its effectiveness and robustness. Previous works use the replacement-based backdoor attack for graph classification yet no existing study has examined how replacement-based attack scales up when multiple triggers needs to coexist. In this study, we investigate both existing state of the art replacement-based backdoor attack and our proposed \texttt{BadImplant} attack across five major datasets in multi-targeted attack scenarios for the comparison. The model architecture and training parameters are identically configured for both approaches. The evaluation results are summarized in Table \ref{table:attack_datasets}, where our proposed attack shows its capability of compromising the GNN model while conventional replacement-based attack fails to support the multi-targeted attack.

As shown in those datasets rich with node features (MNIST, CIFAR-10, ENZYMES), our proposed attack has achieved exceptional performance, while the replacement-based approach displayed severe limitations. On MNIST dataset, the result demonstrates that the subgraph replacement-based attack failed to implant three triggers simultaneously achieving only 2.5\%, 21.90\% and 78.12\% on ASR, while our proposed attack can attain ASRs with the minimal of 99.81\%. In case of attack evasiveness, subgraph replacement attack degrades the clean accuracy to only 18.06\% from the clean model accuracy of 90.60\% whereas our approach attain the clean accuracy of 90.50\%. ENZYMES dataset evaluation is also consistent that the subgraph replacement attack achieves ASRs of 19.8\%, 14\%, 75.49\% with clean accuracy drop (CAD) of 34.17\%, while our proposed attack achieves ASRs in range of 89\% to 94\% with CAD of merely 3.34\%. The relatively lower performance on ENZYMES dataset compared to MNIST evaluation is due to its smaller training graph data which is 100 times lower than MNIST. Besides, the subgraph replacement-based attack significantly degrades the clean accuracy, leading to the inapplicability of trained model. Note that the only dataset where the replacement-based attack was successful for implementing multi-targeted attack is on CIFAR-10 where it achieves over 99\% ASRs with only 0.89\% CAD. Yet, our proposed mechanism still outperforms with ASRs above 99.95\%, 99.89\% and 99.94\% and CAD 0.75\%.

For those datasets with only topology (Reddit-Multi-12k and Reddit-Multi-5k) where the classification depends solely on the geometry of the graph, implanting multiple triggers into the model becomes a nontrivial task, as it drastically increases risks of dominant trigger interfering with the models learning process of other triggers. The result on Table \ref{table:attack_datasets} demonstrates that our proposed attack continues outperforming the conventional approach by large margins. Conventional replacement-based method exhibits a small CAD of 0.92\% on Reddit-Multi-12k dataset but only achieves ASRs of 58.49\%, 87\% and 57.14\%, indicating the attack incapability. In contrast, our proposed approach achieves ASRs of minimal 98.72\% with the maximal CAD of 0.5\%. The comparable results can be observed on Reddit-Multi-5k dataset.
% we attained similar kind of results where the subgraph replacement achieved only one ASR above 82\% with a very small CAD of 0.1\%. In comparison, our proposed attack mechanism achieved an ASRs ranging from 96\% to 99.25\% with a CAD of 0.3\%.  

To this end, our proposed multi target backdoor attack demonstrate exceptional performance by successfully embedding multiple triggers into the GNN model without the interference among the triggers while preserving the clean accuracy of the model. The main results in Table \ref{table:attack_datasets} empirically validate the clear superiority of our proposed attack over the conventional replacement-based attack from all aspects in case of multi-targeted backdoor attack on graph classification.

\begin{table}[h]
\centering
\caption{\small Attack performance on different models}
\renewcommand{\arraystretch}{1.3}
\resizebox{\columnwidth}{!}{%
\begin{tabular}{c|ccc|ccc}
\hline
\textbf{Model} &
  \textbf{ASR\_1} &
  \textbf{ASR\_2} &
  \textbf{ASR\_3} &
  \textbf{CA} &
  \textbf{\begin{tabular}[c]{@{}c@{}}Clean Model \\ Accuracy\end{tabular}} &
  \textbf{CAD} \\ \hline
GCN       & 99.95\%                        & 99.89\% & 99.94\% & 51.81\% & 52.56\% & 0.75\% \\ \hline
GAT       & {\color[HTML]{000000} 99.90\%} & 99.95\% & 99.86\% & 60.35\% & 62.05\% & 1.70\% \\ \hline
GraphSAGE & 99.90\%                        & 99.77\% & 99.73\% & 66.80\% & 68.04\% & 1.24\% \\ \hline
GIN       & 100\%                          & 100\%   & 100\%   & 55.04\% & 57.83\% & 2.79\% \\ \hline
\end{tabular}%
}

\label{table:attack_models}
\end{table}

\subsubsection{Ablation Evaluations}

To evaluate the robustness of our proposed backdoor attack, we extend our study on additional experiments via the various settings of model architectures, parameters as well as attack configurations. 

\vspace*{2.5mm}

\noindent \textbf{\textit{Model-Agnostic Attack Evaluation: }} To investigate if our multi target backdoor attack generalizes across different GNN models without having any knowledge about the model architecture, model parameters and training parameters, we evaluate our attack against four different widely used baseline GNN models (GCN, GAT, GraphSAGE, GIN) using CIFAR-10 dataset. The results are presented in Table \ref{table:attack_models} which show that our multi-targeted backdoor attack has consistently achieved ASRs greater than 99.7\% on all four GNN models and nearly perfect 100\% ASR on GIN model architecture. In case of clean accuracy, it remains modest: 0.75\% (GCN), 1.70\% (GAT), 1.24\% (GraphSAGE), and 2.79\% (GIN). For implementing our attack, the attack design parameter and approach remain constant for all the models. In Table \ref{table:attack_models}, it can be observed that our multi-targeted attack performs exceptionally regardless of the GNN model architecture, model parameters and training parameters being adopted which validates the model agnostic characteristic behavior of our proposed backdoor attack.

\vspace*{2.5mm}

\noindent \textbf{\textit{Impact of Trigger Injection Methods: }}We evaluate the effect of different injection strategies of the trigger on Reddit-Multi-12k dataset. We test the influence of five distinct injection strategy - highest degree, highest similarity, lowest similarity, lowest degree and random. Reddit-Multi-12k is chosen for this experiment because of its structural variation of the graphs in which has nodes with inconsistent degrees and connections. The comparison of the ASRs and CADs for different injection strategies examined in this study is illustrated in Figure \ref{fig:injection_method}. Deterministic strategies (highest degree, highest similarity, lowest similarity and lowest degree) have achieved ASRs in a range of 97.75\% to 99.42\% with clean accuracy degradation in between 2.46\% and 2.96\%. Although all four of these have attained notable performance, the best performance is achieved by random injection strategy in case of both model manipulation (ASR\(>\)99.89\%) and attack stealthiness (CAD 0.75\%). It suggests that since the triggers using deterministic methods are injected consistently in specific type of node, the backdoor information might interfere with the property of the host graph which decrease model's ability to learn clean graphs. Random injection avoid that by distributing the trigger on different nodes. In addition, the randomness of the trigger injection makes it difficult to detect the trigger by the defender. Based on the evaluated results, we adopt the random injection method for implementing \texttt{BadImplant}. 

\begin{center}
   \captionsetup{type=figure}
   \begin{subfigure}[b]{0.75\linewidth}
       \centering
       \includegraphics[width=\linewidth]{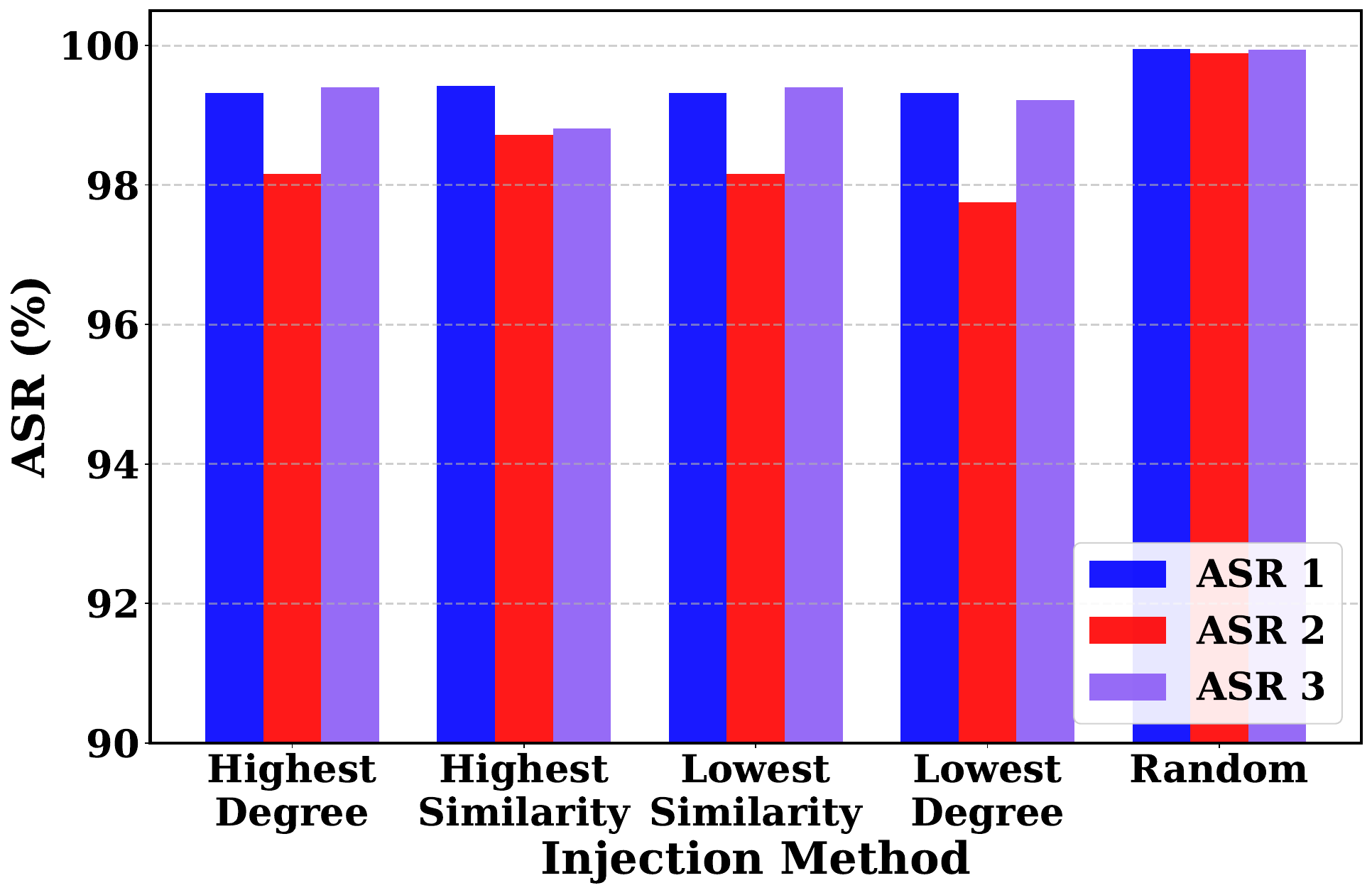}
       \caption{}
       \label{fig:injection_asr}
   \end{subfigure}
   \hfill
   \begin{subfigure}[b]{0.75\linewidth}
       \centering
       \includegraphics[width=\linewidth]{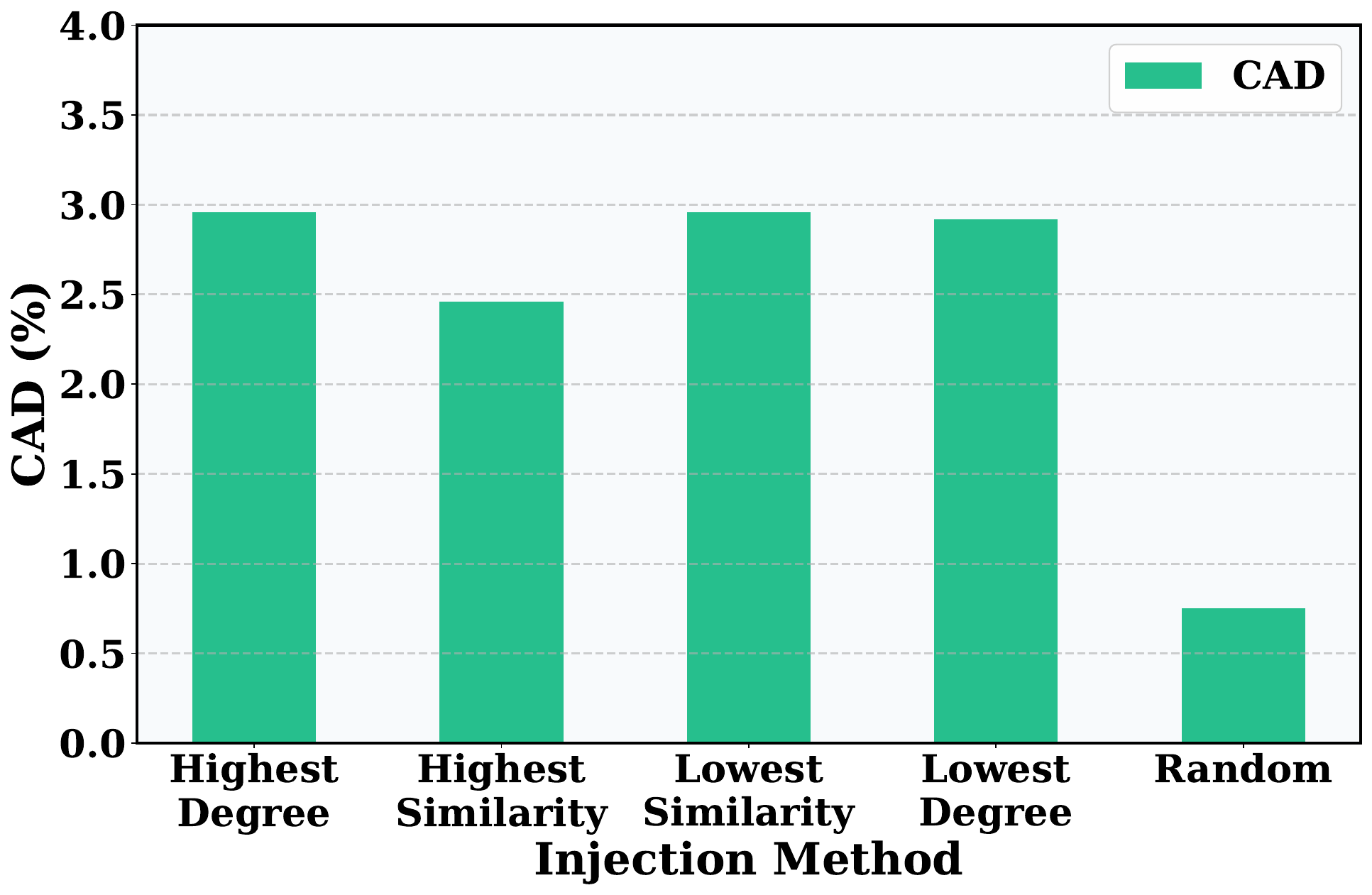}
       \caption{}
       \label{fig:injection_cad}
   \end{subfigure}
   \captionof{figure}{Impact of trigger injection method on (a) ASR and (b) CAD}
   \label{fig:injection_method}
\end{center}

% The combination of superior ASR, minimal CAD and enhanced resistance to statistical detection makes random injection the optimal choice across diverse attack scenarios. 

% \begin{table}[ht]
% \centering
% \resizebox{\columnwidth}{!}{%
% \begin{tabular}{c|rrr|rr}
% \hline
% \textbf{Injection Method} &
%   \multicolumn{1}{c}{\textbf{ASR\_1}} &
%   \multicolumn{1}{c}{\textbf{ASR\_2}} &
%   \multicolumn{1}{c}{\textbf{ASR\_3}} &
%   \multicolumn{1}{c}{\textbf{CA}} &
%   \multicolumn{1}{c}{\textbf{CAD}} \\ \hline
% Clean model &
%   \multicolumn{1}{c}{-} &
%   \multicolumn{1}{c}{-} &
%   \multicolumn{1}{c}{-} &
%   \multicolumn{1}{c}{52.56\%} &
%   \multicolumn{1}{c}{-} \\ \hline
% Highest Degree    & {\color[HTML]{000000} 99.32\%} & 98.16\% & 99.40\% & 49.60\% & 2.96\% \\ \hline
% Higest Similarily & 99.42\%                        & 98.72\% & 98.81\% & 50.10\% & 2.46\% \\ \hline
% Lowest Similarity & 99.32\%                        & 98.16\% & 99.40\% & 49.60\% & 2.96\% \\ \hline
% Lowest Degree     & 99.32\%                        & 97.75\% & 99.22\% & 49.64\% & 2.92\% \\ \hline
% Random            & 99.95\%                        & 99.89\% & 99.94\% & 51.81\% & 0.75\%
% \end{tabular}%
% }
% \caption{\small Impact of injection method on Reddit-Multi-12k}
% \label{table:attack_injection_method}
% \end{table}

\vspace*{2.5mm}

\noindent \textbf{\textit{Impact of Number of Connections: }}To examine the effect of different number of edges and connections (\(k\)) between the trigger subgraph and target graph, number of connection are ranged from one to five summarized in Table \ref{table:attack_number_connection}. Our attack was significantly effective, achieving nearly perfect close to 100\% ASR with a slight drop of clean accuracy below 1.07\% across all configurations. For instance, with the single connection (k=1), \texttt{BadImplant} achieves ASRs of 99.95\%, 99.89\% and 99.94\% with a 0.75\% CAD, demonstrating that minimal integration is sufficient for implementing highly effective backdoor implantation.

\begin{table}[h]
\renewcommand{\arraystretch}{1.15}
\centering
\caption{\small Impact of number of connections}
\resizebox{\columnwidth}{!}{%
\begin{tabular}{c|ccc|cc}
\hline
\textbf{\begin{tabular}[c]{@{}c@{}}Number of \\ Connection\end{tabular}} & \textbf{ASR\_1} & \textbf{ASR\_2} & \textbf{ASR\_3} & \textbf{CA} & \textbf{CAD} \\ \hline
Clean model & -                              & -       & -       & 52.56\% & -      \\ \hline
1           & 99.95\%                        & 99.89\% & 99.94\% & 51.81\% & 0.75\% \\ \hline
2           & {\color[HTML]{000000} 99.98\%} & 99.91\% & 99.95\% & 51.49\% & 1.07\% \\ \hline
3           & 99.92\%                        & 99.92\% & 99.91\% & 51.57\% & 0.99\% \\ \hline
4           & 99.95\%                        & 99.90\% & 99.86\% & 51.81\% & 0.75\% \\ \hline
5           & 99.98\%                        & 99.87\% & 99.94\% & 51.75\% & 0.81\% \\ \hline
\end{tabular}%
}

\label{table:attack_number_connection}
\end{table}

\begin{figure*}[h!]
    \centering
    \begin{subfigure}[b]{0.29\textwidth}
        \centering
        \includegraphics[width=\textwidth]{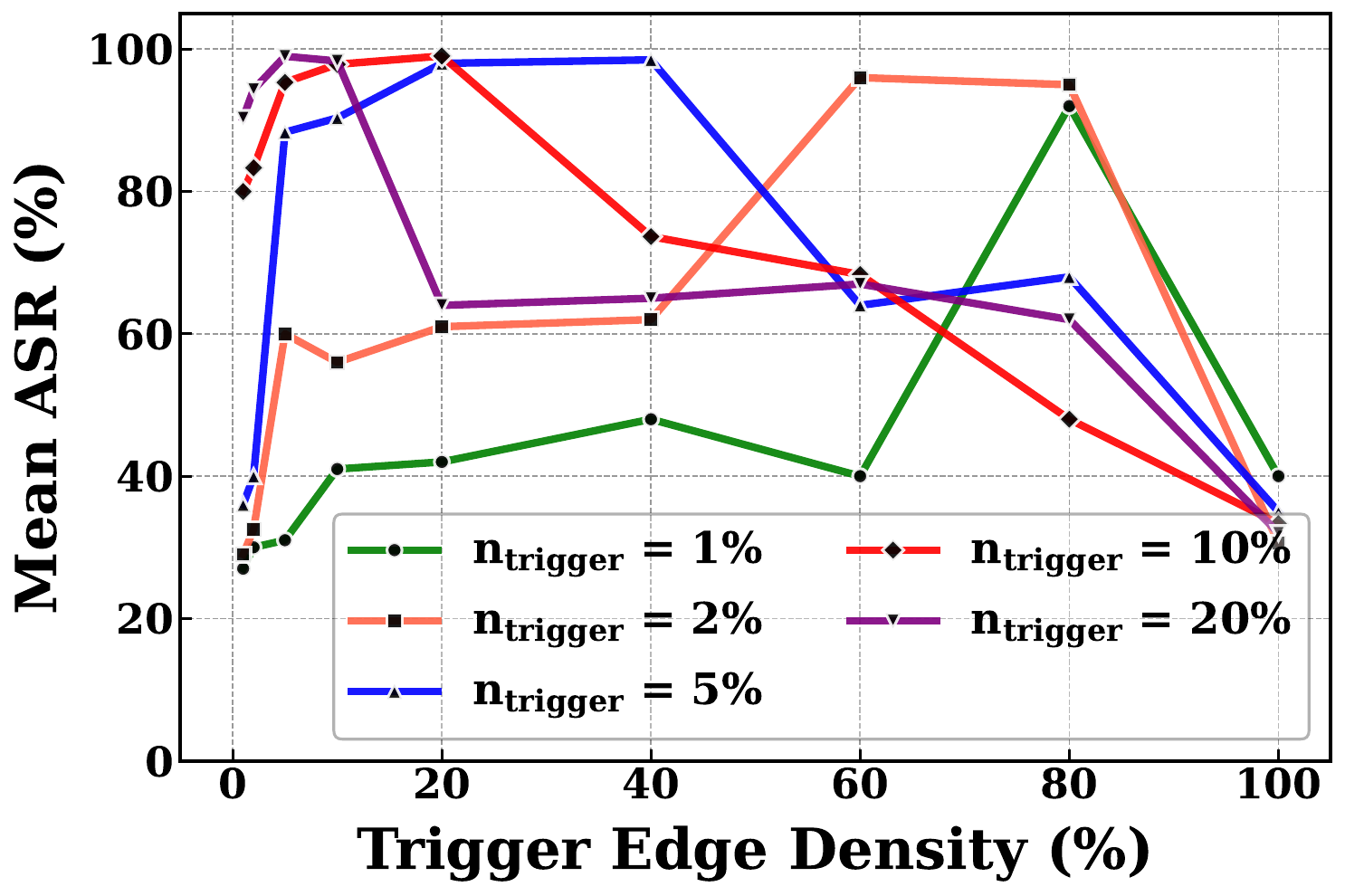}
        \caption{\small Mean ASR for different \(n_{trigger}\) \& \(n_{density}\) settings}
        \label{fig:trig_size_density}
    \end{subfigure}
    \hfill
    \begin{subfigure}[b]{0.4\textwidth}
        \centering
        \includegraphics[width=\textwidth]{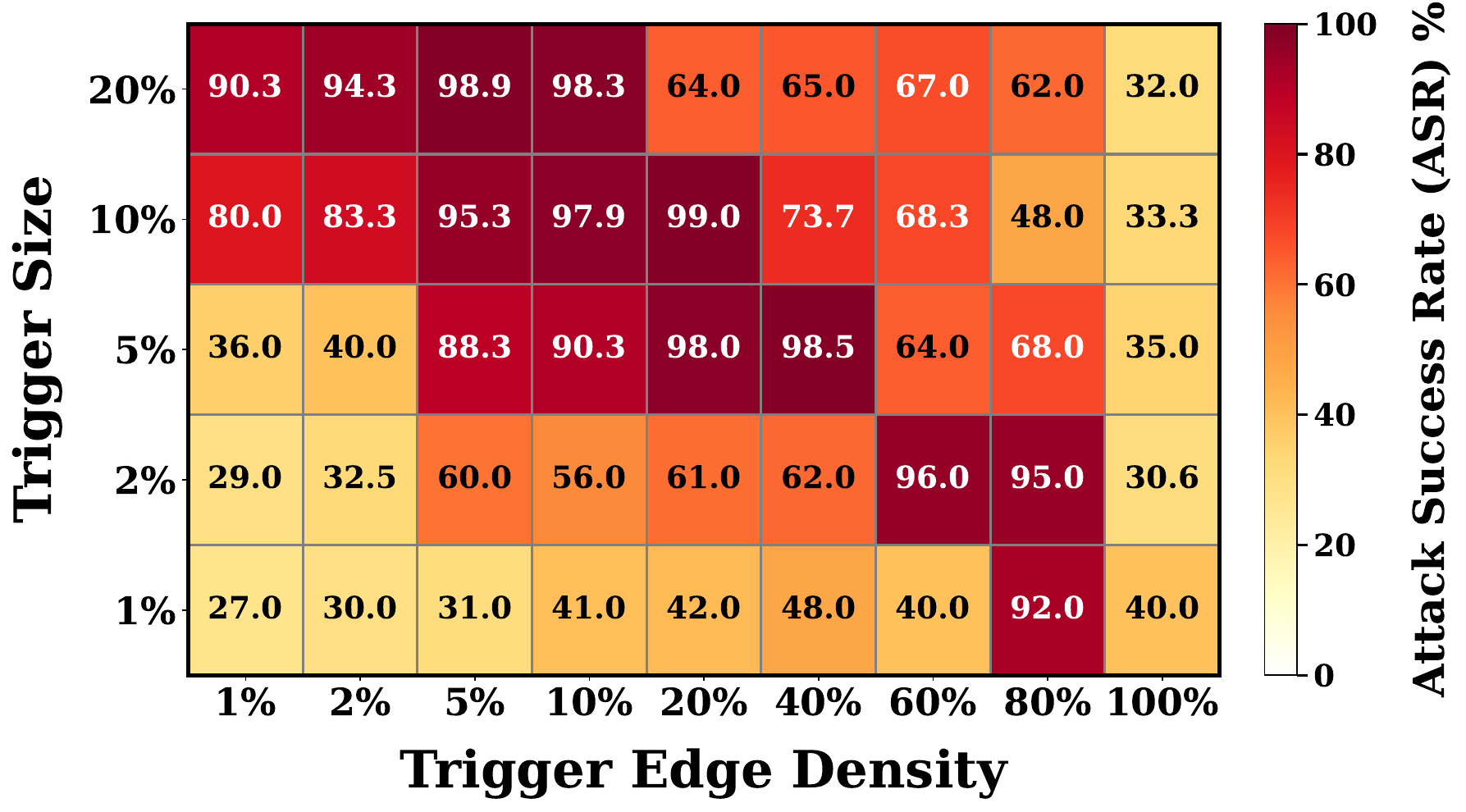}
        \caption{\small Heat map for \(n_{trigger}\) \& \(n_{density}\) settings}
        \label{fig:heat_map}
    \end{subfigure}
    \hfill
    \begin{subfigure}[b]{0.29\textwidth}
        \centering
        \includegraphics[width=\textwidth]{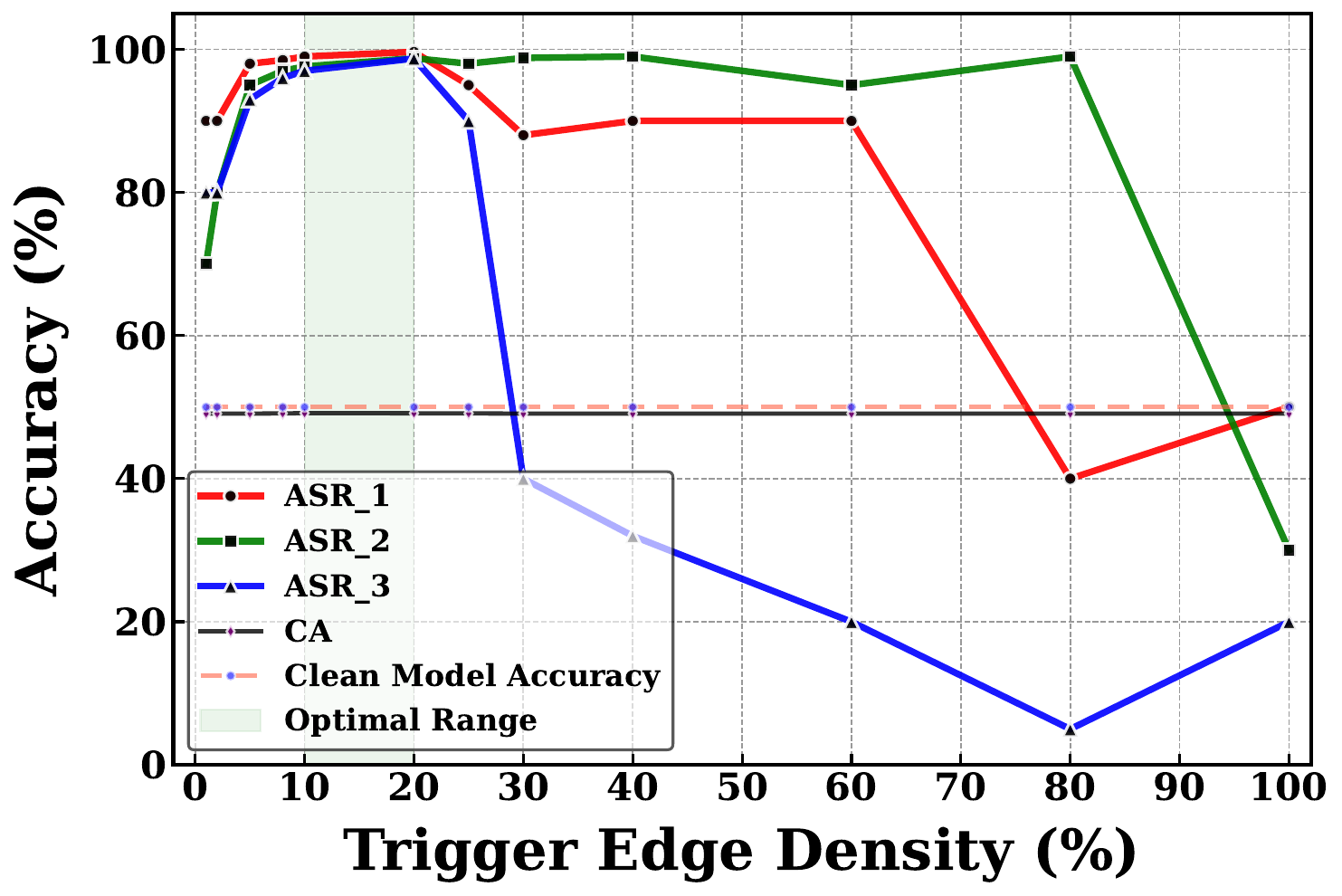}
        \caption{\small Impact of trigger edge density \(n_{trigger} = 10\%\)}
        \label{fig:trig_density}
    \end{subfigure}
    \caption{Analysis on Reddit-Multi-12k dataset}
    \label{fig:combined_analysis}
\end{figure*}

\vspace*{2.5mm}

\noindent \textbf{\textit{Scalability to Multiple Targets: }} We have used number of target labels as three as default for all the experiments conducted in this study. A fundamental question for our multi-targeted attack arises regarding its scalability: if we increase the number of target to a higher number from 3, can \texttt{BadImplant} support the coexistence of those triggers without any interference to the GNN model and the clean accuracy degradation? To investigate such question, we increase the number of target from 3 to the maximum (total number of classes in a given dataset). Table \ref{table:attack_targets} shows the attack success rate (ASR), clean accuracy (CA) and clean accuracy drop (CAD) of \texttt{BadImplant} for various number of target classes. The ASR is the mean value of all ASRs for each target class. With our default settings of three-targeted attack, \texttt{BadImplant} achieves a mean ASR of 99.93\% without any meaningful drop of clean accuracy (approx. 0.75\%). Such trend continues when we increase the number of target to four and five. Even when the number of target is increased to 8 and 10 (which is the maximum number of class in CIFAR-10 dataset), we can still achieve mean ASR of 99.87\%. Such result demonstrate that trigger design  can coexist without any interference with our proposed framework. The CAD increased to 3.3\% when number of trigger is increased to 10, because for implanting larger number of triggers, greater number of graphs are needed to be poisoned, which subsequently increases the share of poisoned graph against clean graph for backdooring the model. Clean accuracy drop is considered acceptable (under 3\%) despite the model is poisoned by larger number of triggers.

{
\setlength{\tabcolsep}{11pt}
\begin{table}[ht]
\renewcommand{\arraystretch}{1.15}
\centering
\caption{\small Results for backdoor attack on different number of target class}
\begin{tabular}{c|r|r|r}
\hline
\multicolumn{1}{c}{\textbf{\#Target}} & \multicolumn{1}{c}{\textbf{ASR}} & \multicolumn{1}{c}{\textbf{CA}} & \multicolumn{1}{c}{\textbf{CAD}} \\ \hline
\multicolumn{1}{l}{Clean model} & \multicolumn{1}{c}{-}            & \multicolumn{1}{l}{52.56\%}     & \multicolumn{1}{c}{-}            \\ \hline
3  & 99.93\%                        & 51.81\% & 0.75\% \\ \hline
4  & {\color[HTML]{000000} 99.89\%} & 51.36\% & 1.20\% \\ \hline
5  & 99.92\%                        & 51.38\% & 1.18\% \\ \hline
8  & 99.89\%                        & 50.10\% & 2.46\% \\ \hline
10 & 99.87\%                        & 49.43\% & 3.13\% \\ \hline
\end{tabular}%

\label{table:attack_targets}
\end{table}
}

\vspace*{2.5mm}

\noindent \textbf{\textit{Effect of Poisoning Ratio, Trigger Size \& Edge Density: }} To fully understand how the attack design parameters (poisoning ratio, trigger size \& trigger edge density) influence the performance, we vary them once at a time while fixing all the other configurable parameters and flexible options as our default settings. 
% These parameters fundamentally maintain the balance between the effectiveness and stealthiness, with optimum value of these parameters depends on the dataset characteristic. 

For evaluating the impact of these parameters on dataset with node features (\(1^{st}\) type datasets), we evaluate these parameters on CIFAR-10 dataset, where the prediction depends on global level embeddings which is a combination of all the available embeddings (node features, edge feature, graph level feature) and connections among the nodes. Table \ref{table:attack_poisoning_ratio}, \ref{table:attack_trigger_size} and \ref{table:attack_trigger_edge_density} summarize the impact of poisoning ratio, trigger size and edge density respectively on CIFAR-10 dataset where each parameter is varied individually while all other configurable parameters are set at the default settings. We achieve significantly high ASRs for a wide range of all of these parameters. The ASR of 1\% poisoning ratio is above 99\%, while it increases to a 100\% ASR for poisoning ratio at 30\%. The downside of increased poison ratio is the drop of model clean accuracy from 52.04\% to 41.23\%.

\begin{table}[ht]
\renewcommand{\arraystretch}{1.15}
\centering
\caption{\small Impact of poisoning ratio}
\resizebox{\columnwidth}{!}{%
\begin{tabular}{c|ccc|cc}
\hline
\textbf{Poisoning Ratio} & \textbf{ASR\_1}                & \textbf{ASR\_2} & \textbf{ASR\_3} & \textbf{CA} & \textbf{CAD} \\ \hline
Clean model & -        & -       & -       & 52.56\% & -       \\ \hline
1\%         & 99.86\%  & 99.72\% & 99.71\% & 52.04\% & 0.52\%  \\ \hline
2\%                   & {\color[HTML]{000000} 99.92\%} & 99.84\%         & 99.80\%         & 52.39\%     & 0.17\%       \\ \hline
5\%         & 99.95\%  & 99.89\% & 99.94\% & 51.81\% & 0.75\%  \\ \hline
10\%        & 99.93\%  & 99.95\% & 99.95\% & 51.17\% & 1.39\%  \\ \hline
15\%        & 99.98\%  & 99.95\% & 99.93\% & 49.90\% & 2.66\%  \\ \hline
20\%        & 100.00\% & 99.98\% & 99.95\% & 48.03\% & 4.53\%  \\ \hline
30\%        & 100\%    & 100\%   & 100\%   & 41.23\% & 11.33\% \\ \hline
\end{tabular}%
}

\label{table:attack_poisoning_ratio}
\end{table}

\begin{table}[ht]
\renewcommand{\arraystretch}{1.15}
\centering
\caption{\small Impact of trigger size}
\resizebox{\columnwidth}{!}{%
\begin{tabular}{c|ccc|cc}
\hline
\textbf{Trigger Size} & \textbf{ASR\_1}                & \textbf{ASR\_2} & \textbf{ASR\_3} & \textbf{CA} & \textbf{CAD} \\ \hline
Clean model & -        & -       & -       & 52.56\% & -      \\ \hline
2\%         & 99.54\%  & 99.28\% & 99.00\% & 52.17\% & 0.39\% \\ \hline
5\%                   & {\color[HTML]{000000} 99.68\%} & 99.64\%         & 99.60\%         & 51.60\%     & 0.96\%       \\ \hline
10\%        & 99.92\%  & 99.86\% & 99.92\% & 52.12\% & 0.44\% \\ \hline
15\%        & 99.94\%  & 99.86\% & 99.96\% & 52.11\% & 0.45\% \\ \hline
20\%        & 99.95\%  & 99.89\% & 99.94\% & 51.81\% & 0.75\% \\ \hline
25\%        & 99.95\%  & 99.93\% & 99.95\% & 51.76\% & 0.80\% \\ \hline
30\%        & 100.00\% & 99.94\% & 99.94\% & 52.15\% & 0.41\% \\ \hline
\end{tabular}%
}

\label{table:attack_trigger_size}
\end{table}

\begin{table}[ht]
\renewcommand{\arraystretch}{1.2}
\centering
\caption{\small Impact of trigger edge density}
\resizebox{\columnwidth}{!}{%
\begin{tabular}{c|ccc|cc}
\hline
\textbf{Trigger Edge Density} & \textbf{ASR\_1}                & \textbf{ASR\_2} & \textbf{ASR\_3} & \textbf{CA} & \textbf{CAD} \\ \hline
Clean model & -       & -       & -       & 52.56\% & -      \\ \hline
20\%        & 99.88\% & 99.84\% & 99.81\% & 51.97\% & 0.59\% \\ \hline
40\%                  & {\color[HTML]{000000} 99.94\%} & 99.74\%         & 99.86\%         & 51.34\%     & 1.22\%       \\ \hline
60\%        & 99.94\% & 99.94\% & 99.94\% & 51.70\% & 0.86\% \\ \hline
80\%        & 99.95\% & 99.89\% & 99.94\% & 51.81\% & 0.75\% \\ \hline
100\%       & 99.92\% & 99.92\% & 99.92\% & 51.57\% & 0.99\% \\ \hline
\end{tabular}%
}

\label{table:attack_trigger_edge_density}
\end{table}

For datasets that lacks node feature matrices containing only topological information (\(2^{nd}\) type dataset), we use degree and node as the node feature for both original graph as well as the trigger design. To this end, the classification exclusively depends on the structure of the graph, which makes setting trigger size and edge density become substantially critical. Unlike (\(1^{st}\)) type of datasets with rich node features that exhibit robustness across parameter ranges, topology-based datasets demand deliberate tuning of trigger size and edge density due to their complex relationship. To comprehensively evaluate the impact of trigger size and edge density, we use different trigger sizes {1\%, 2\%, 5\%, 10\% and 20\%} and vary the trigger edge density from 1\% to 100\% for each of trigger size values. We use Reddit-Multi-12k for evaluation. 

The heatmap in Figure \ref{fig:heat_map} and mean ASR curves in Figure \ref{fig:trig_size_density} exhibit a clear inverse relationship between the trigger size and trigger edge density for a successful attack: larger trigger size require lower trigger edge density for optimal performance, and vice versa. Triggers with 20\% trigger size show better performance (ASR > 95\%) with 2-10\% edge density, while triggers with 5\% size demand 10-20\% density for optimal attack performance. We observe that trigger edge density is inversely proportional to the trigger size for a successful backdoor attack on \(2^{nd}\) type dataset. The heat map shows that \texttt{BadImplant} attains highest mean ASR on 10\% trigger size and 20\% trigger edge density. To visualize how ASRs react to the change of trigger edge density, we plot all the ASRs by varying the trigger edge density with 10\% trigger size. In Figure \ref{fig:trig_density}, we can see that backdoored model with trigger edge density at approximately 10-20\% is able to learn all 3 designed triggers with high ASR between 25\% and 60\%. In summary, our rigorous experimental results show the importance of the attack design parameters and how to set these parameters for different kind of datasets in order to launch a successful backdoor attack.

\begin{figure*}[h!]
\centering
\begin{tabular}{cccc}

    % -------- Row 1 --------
    \begin{subfigure}{0.27\textwidth}
        \centering
        \includegraphics[width=\textwidth]{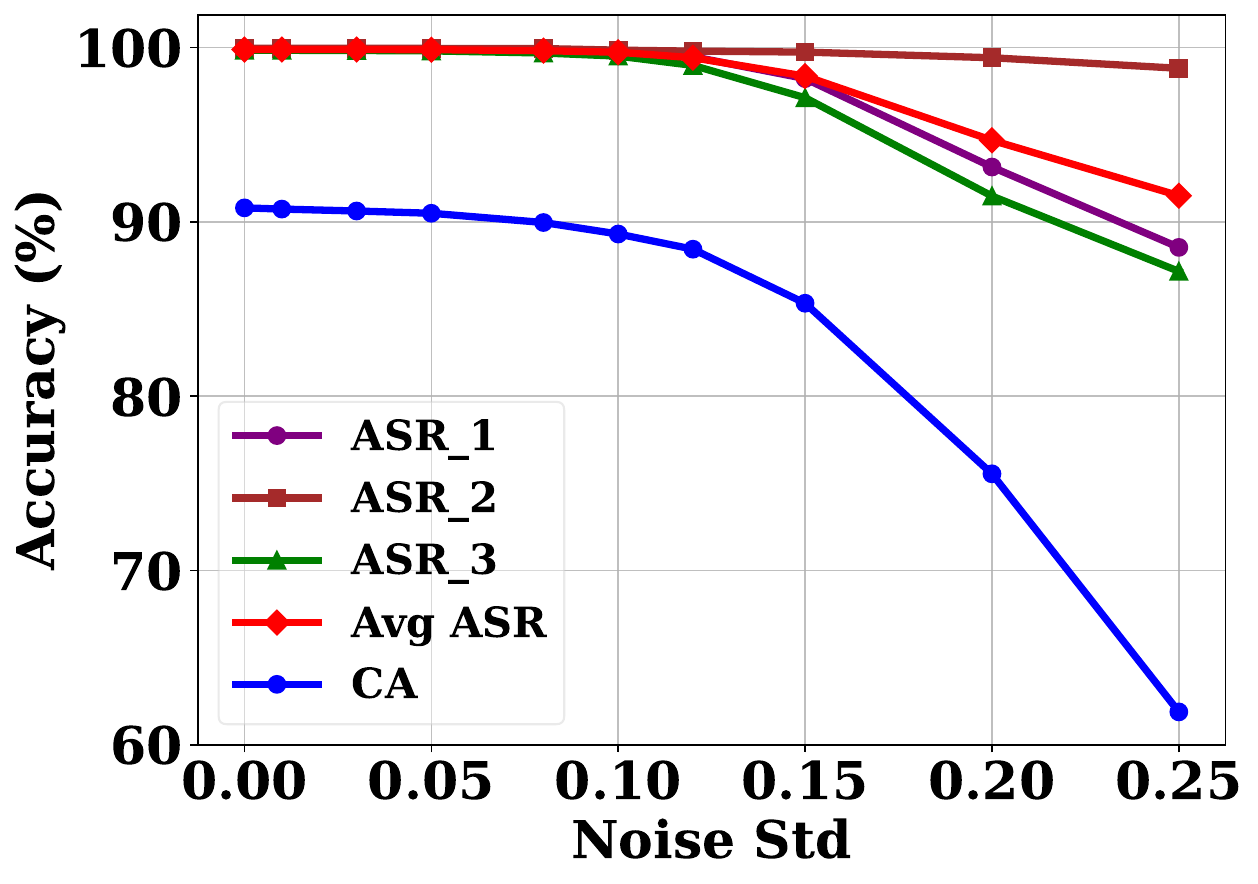}
        \caption{\small MNIST}
        \label{fig:mnist_rs}
    \end{subfigure}
    \hspace*{4mm}

    \begin{subfigure}{0.27\textwidth}
        \centering
        \includegraphics[width=\textwidth]{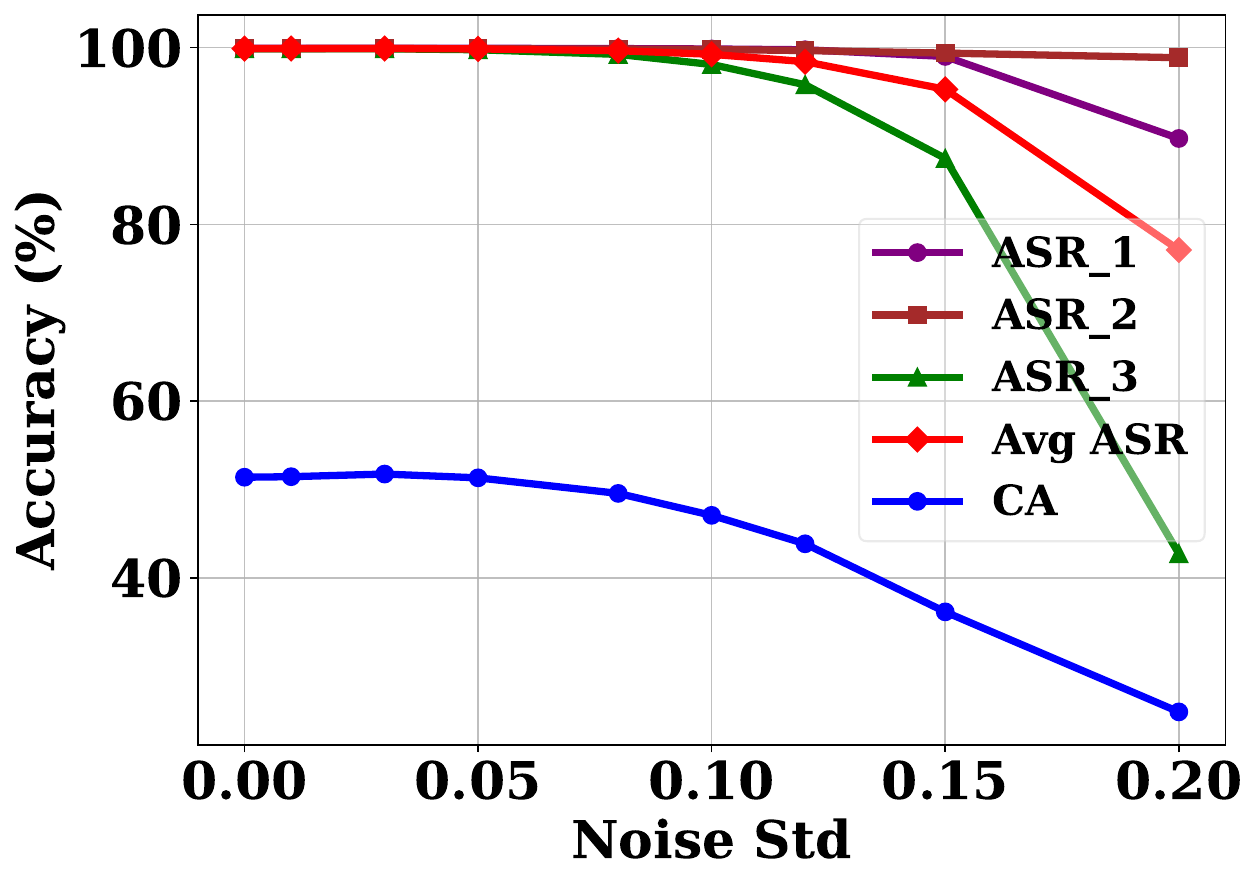}
        \caption{\small CIFAR-10}
        \label{fig:cifar10_rs}
    \end{subfigure} 
    \hspace*{4mm}
    
    \begin{subfigure}{0.27\textwidth}
        \centering
        \includegraphics[width=\textwidth]{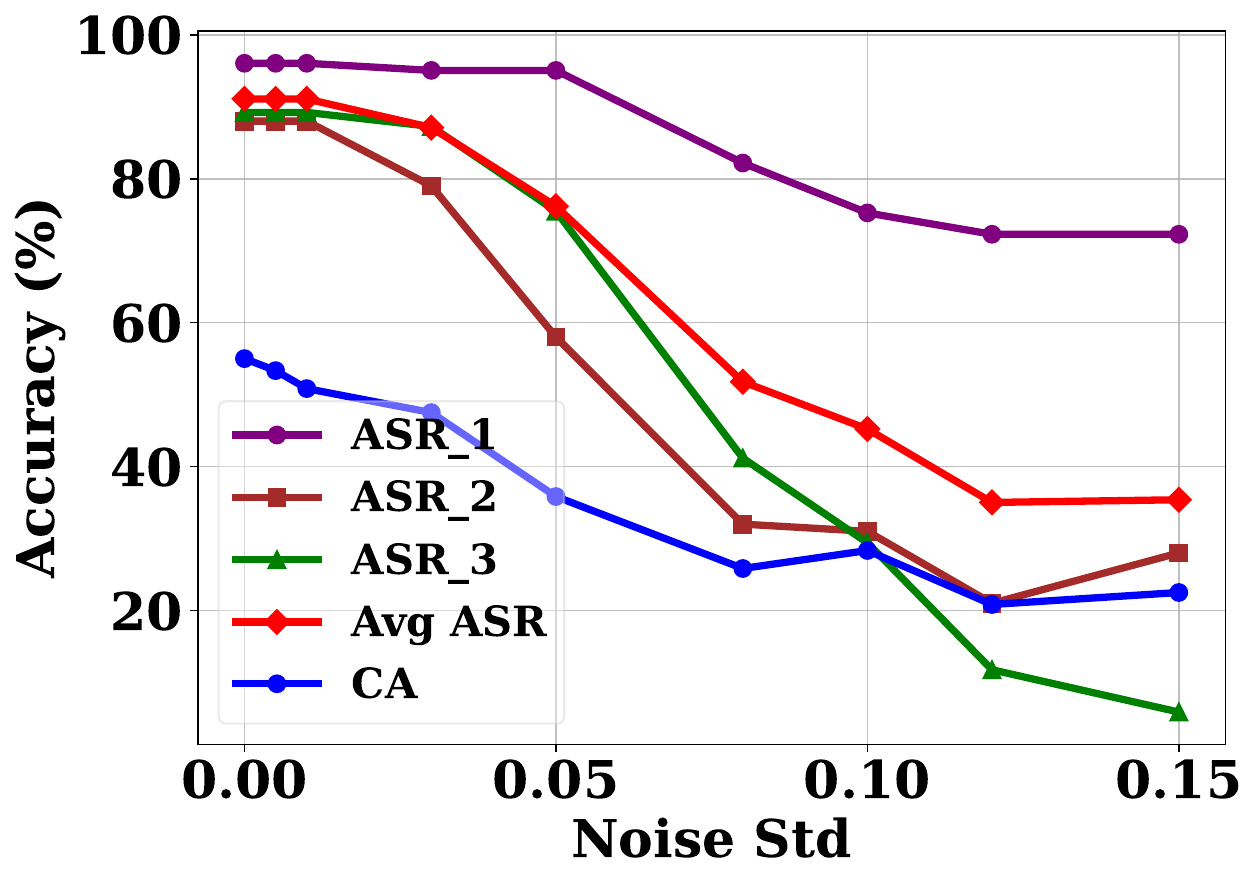}
        \caption{\small ENZYMES}
        \label{fig:RS_ENZYMES}
    \end{subfigure} 
    \\[8pt]

    \begin{subfigure}{0.27\textwidth}
        \centering
        \includegraphics[width=\textwidth]{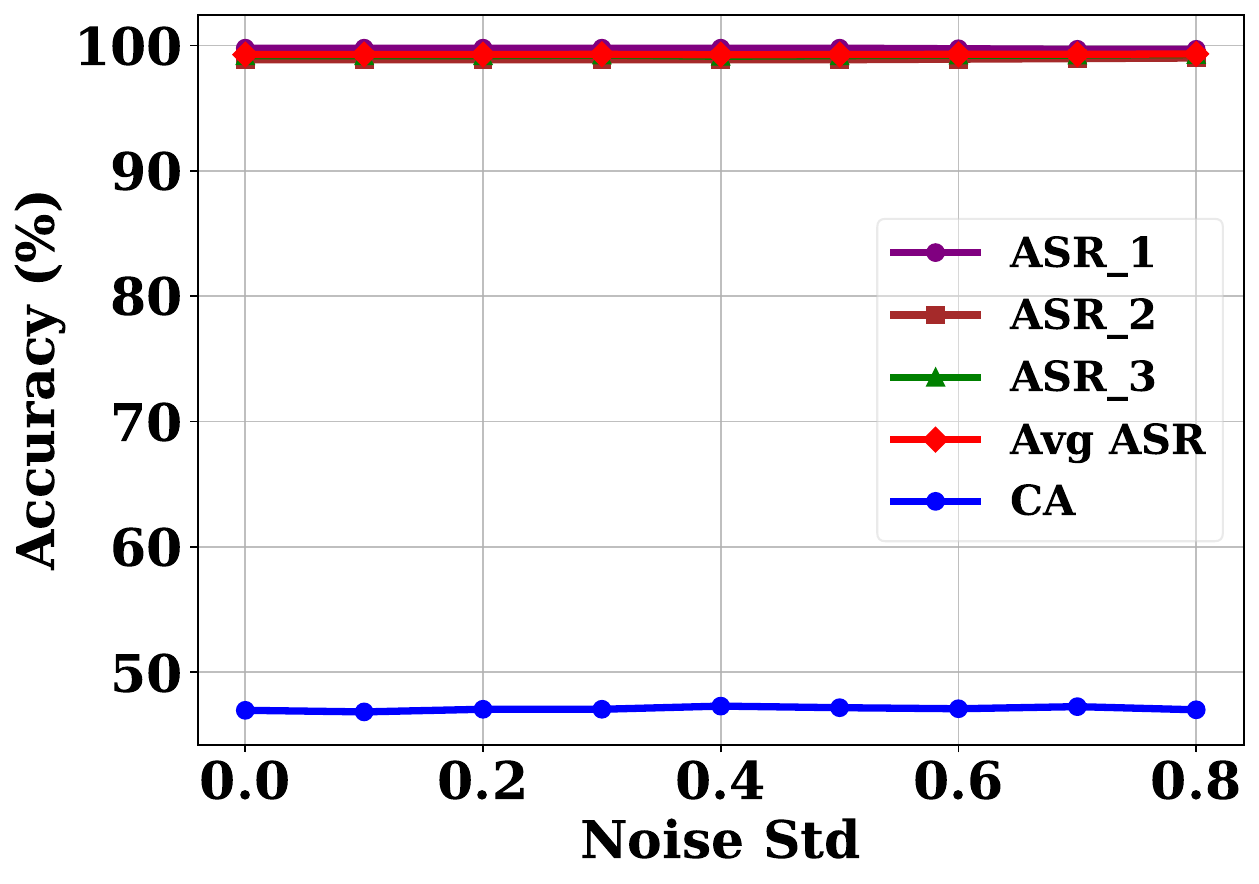}
        \caption{\small Reddit-Multi-12k}
        \label{fig:red_12k_rs}
    \end{subfigure} 
    \hspace*{4mm}
    
    \begin{subfigure}{0.27\textwidth}
        \centering
        \includegraphics[width=\textwidth]{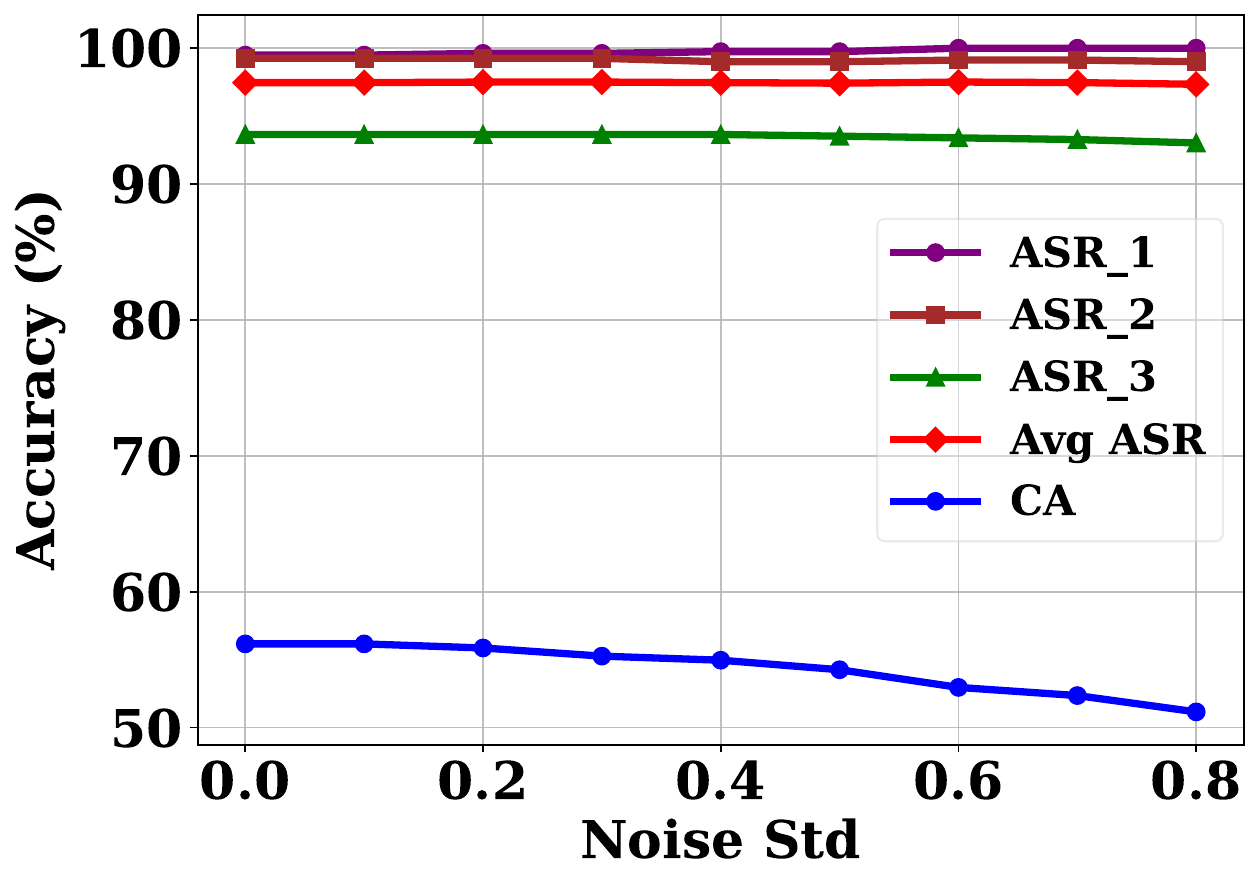}
        \caption{\small Reddit-Muti-5k}
        \label{fig:red_5k_rs}
    \end{subfigure}

\end{tabular}

\caption{\small Evaluation against randomized smoothing (RS) defense on five datasets}
\label{fig:rs_defense_analysis}
\end{figure*}

\begin{figure*}[h!]
\centering
\begin{tabular}{cccc}

    % -------- Row 1 --------
    \begin{subfigure}{0.27\textwidth}
        \centering
        \includegraphics[width=\textwidth]{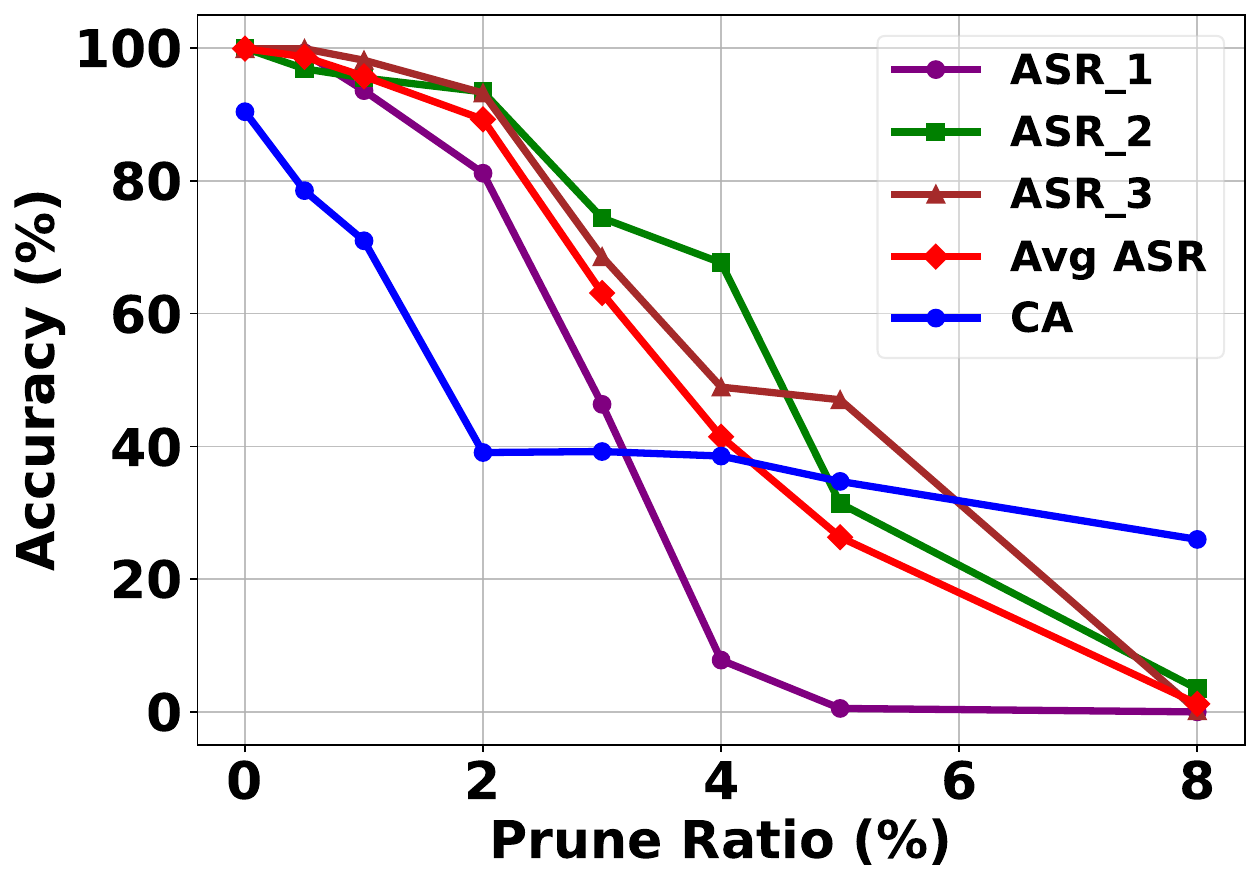}
        \caption{\small MNIST}
        \label{fig:mnist_fp}
    \end{subfigure} 
    \hspace*{4mm}
    
    \begin{subfigure}{0.27\textwidth}
        \centering
        \includegraphics[width=\textwidth]{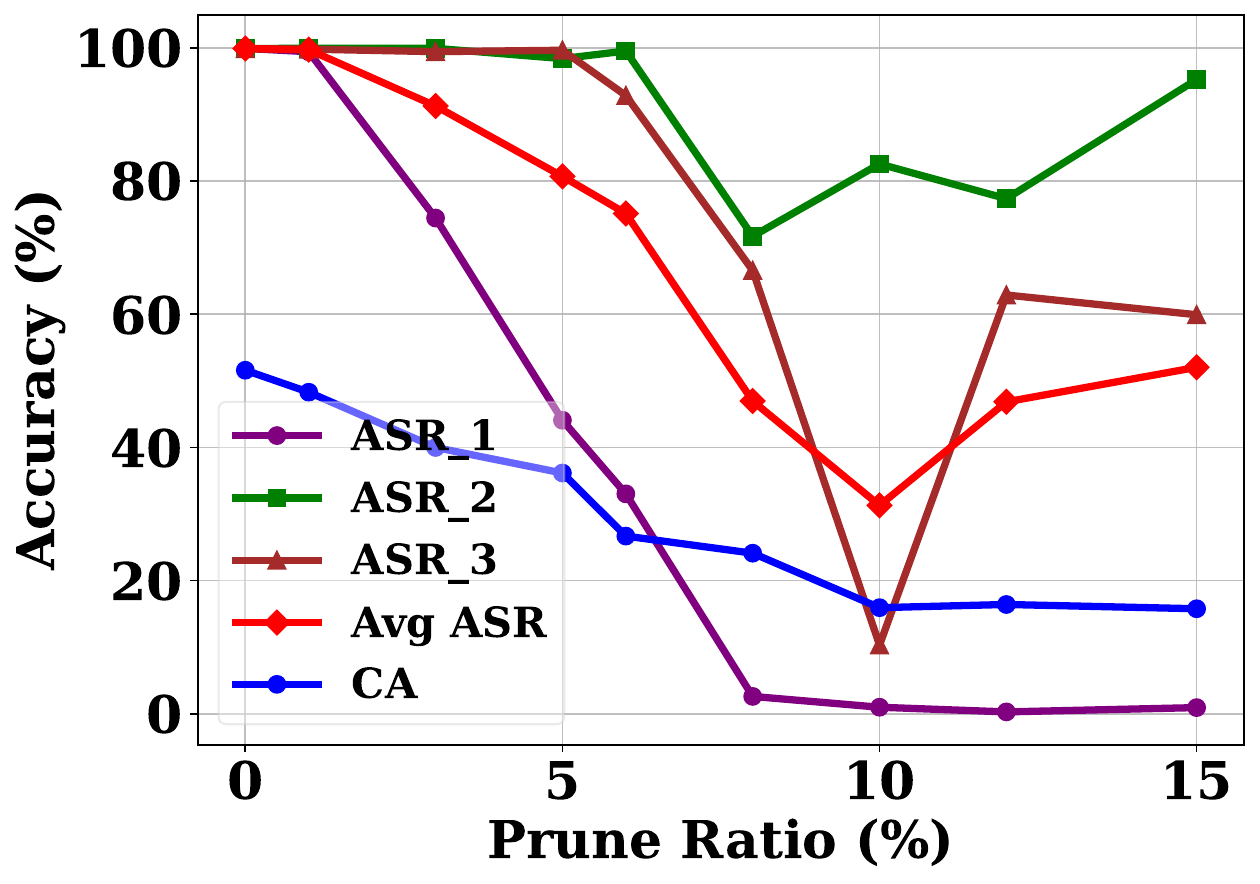}
        \caption{\small Cifar10}
        \label{fig:cifar10_fp}
    \end{subfigure} 
    \hspace*{4mm}
   
    \begin{subfigure}{0.27\textwidth}
        \centering
        \includegraphics[width=\textwidth]{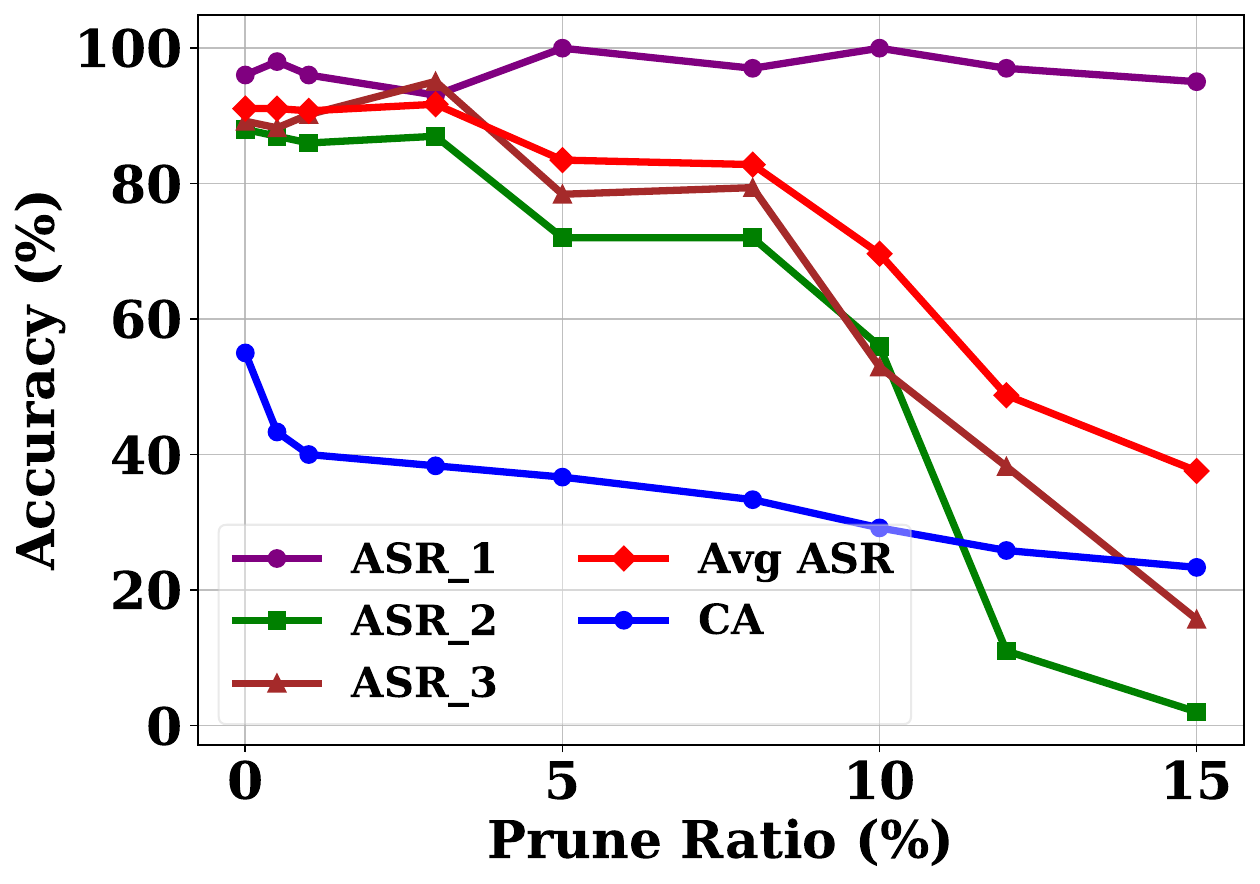}
        \caption{\small ENZYMES}
        \label{fig:FP_ENZYMES}
    \end{subfigure} 
    \\[10pt]

    \begin{subfigure}{0.27\textwidth}
        \centering
        \includegraphics[width=\textwidth]{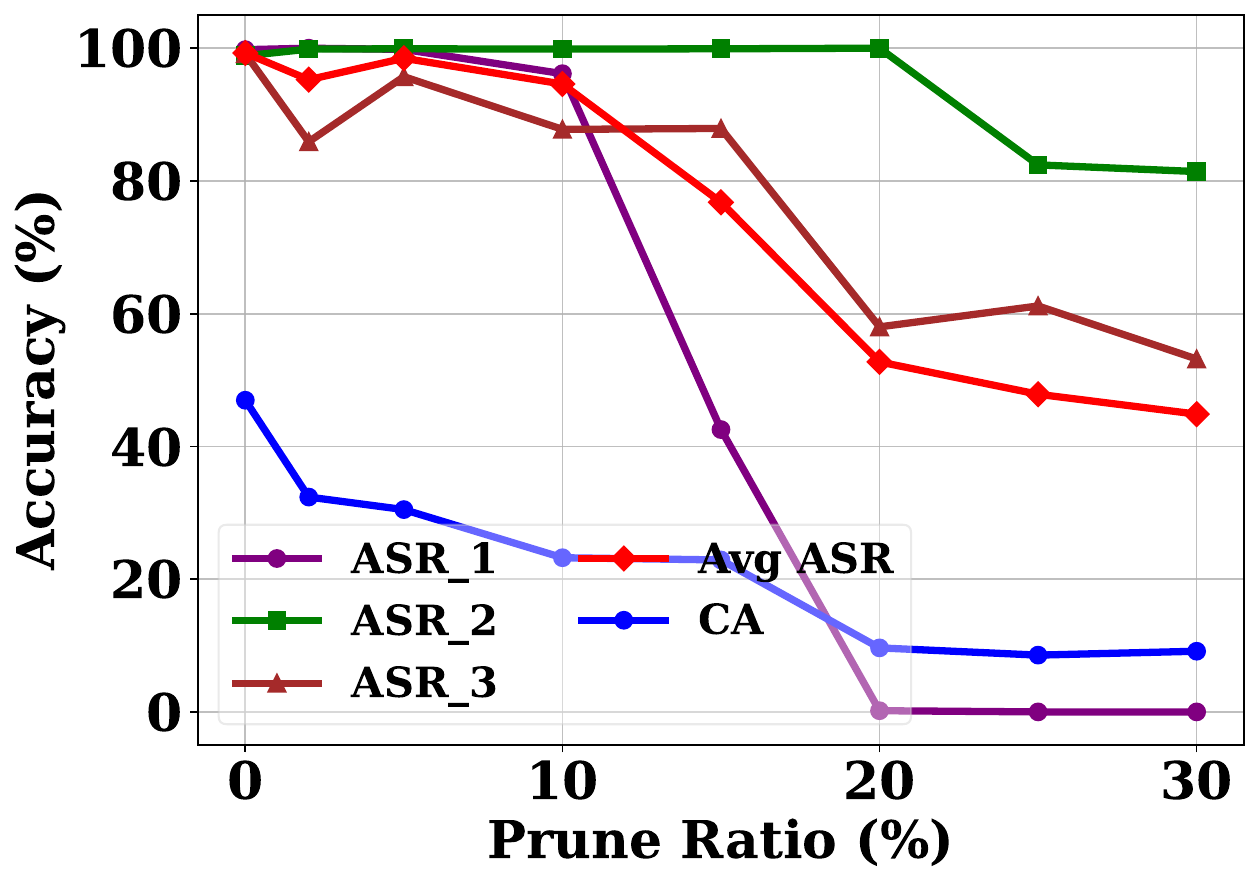}
        \caption{\small Reddit-Multi-12k}
        \label{fig:red_5k_fs}
    \end{subfigure} 
    \hspace*{4mm}
    
    \begin{subfigure}{0.27\textwidth}
        \centering
        \includegraphics[width=\textwidth]{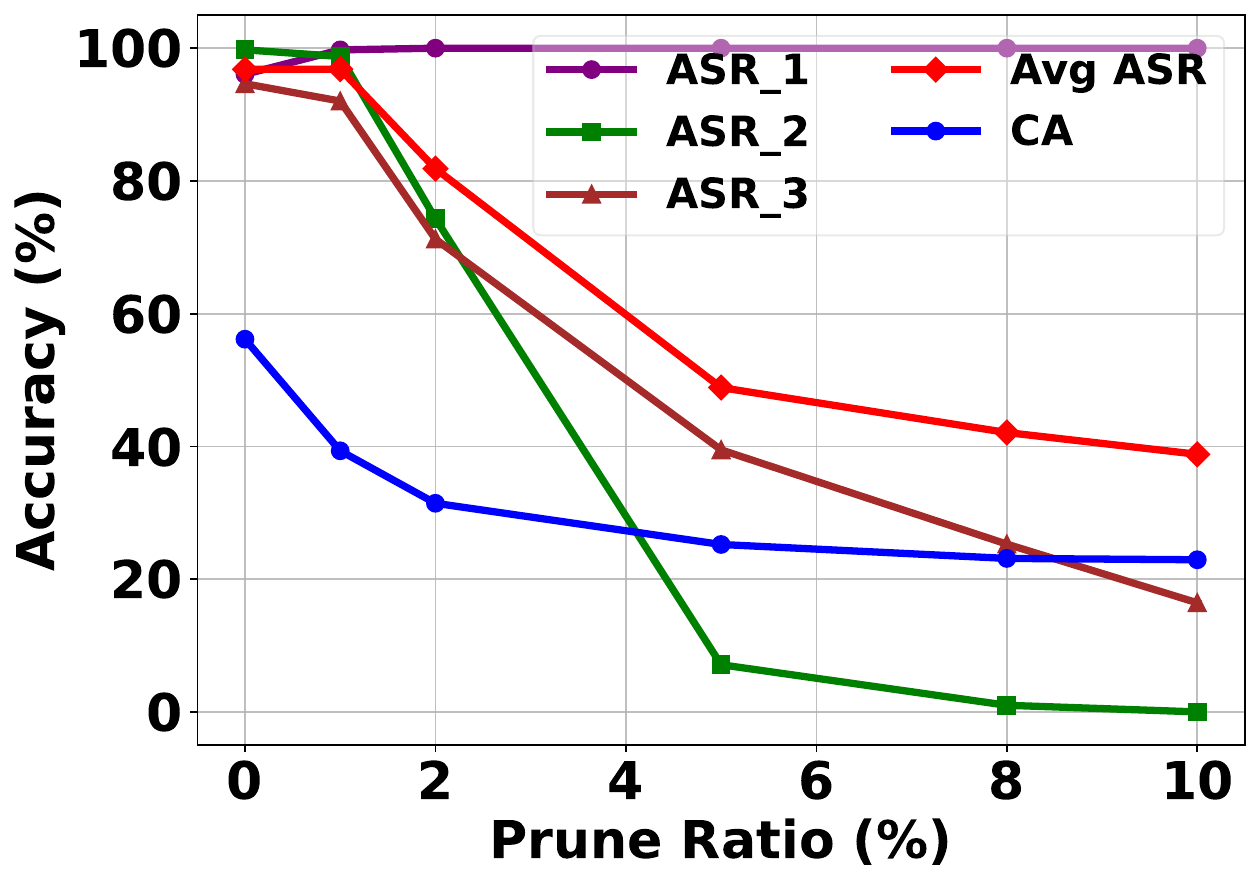}
        \caption{\small Reddit-Multi-5k}
        \label{fig:red_5k_fs}
    \end{subfigure}

\end{tabular}

\caption{\small Evaluation against fine pruning (FP) defense on five datasets}
\label{fig:fp_defense_analysis}
\end{figure*}

\section{Robustness Against Certified Defense}\label{sec:defense}
To evaluate the robustness of our \texttt{BadImplant}, we assess its performance against two widely adopted defense mechanisms: randomized smoothing \cite{cohen2019certified} and fine pruning \cite{liu2018fine}. In this section, we present a brief description of two defense mechanisms, experimental setups and thorough evaluations of \texttt{BadImplant}'s resilience against certified countermeasures.

\subsection{Defense Background}
\noindent \textbf{Randomized Smoothing (RS): } Randomized smoothing is one of the most efficient and popular defense methods against backdoor attacks. Originally proposed by Cohen et al. \cite{cohen2019certified} for certifying adversarial robustness in deep learning classifiers, randomized smoothing constructs a smoothed classifier by introducing stochastic perturbations to the input and aggregating the predictions through majority voting. In the context of GNN backdoor attacks, \cite{xi2021graph} and \cite{zhang2021backdoor} are the first to generalize randomized smoothing as a certified defense for graph classification against graph backdoor attacks. More recently, \cite{wang2025boosting} employs randomized smoothing as one of the defense benchmarks in their evaluation for clean-label backdoor attacks on graph classification. Randomized smoothing adds small gaussian noises \(N(0,\sigma^2I)\) to the node features of the input test graphs and creates variants of the input samples which can be mathematically written as \(G_i^{noisy}\), where i = 1 to n. A smoothed classifier is developed by the prediction on all of these variants and majority voting used to obtain the final prediction:

\[
\hat{y} = \arg\max_{y \in \mathcal{Y}} 
\left| \left\{ i \;:\; f_{\omega}\!\left( G^{\text{noisy}}_{i} \right) = y,\; i = 1,\dots,n \right\} \right|
\]

\noindent where \(G_i^{noisy}\) is the \(i_{th}\) noisy version of the input graph G. Such approach smoothens the decision of the model and makes it more robust to backdoor triggers.

\vspace*{2.5mm}
\noindent \textbf{Fine Pruning (FP): } Fine pruning is a post-training defense that conducts two-step operations on the backdoored model: 1) Fine tune the backdoored GNN model with a fraction of clean graph dataset which modifies the model weights to decrease backdoor activations, and 2) Prune neurons of the GNN model with low activation values on clean graph data. 
% Xi \cite{xi2021graph} evaluates their GTA framework against Neural Cleanase, pruning neurons that activates if data are triggered. Although Neural Cleanase is an effective technique against backdoor attacks, fine pruning is one step ahead of it which combines fine tuning of the model with clean data with neuron pruning techniques. The proportion of pruned neurons is controlled by the pruning ratio. 
\vspace*{0.5mm}

%\noindent \textbf{Edge Pruning (EP): } Edge pruning removes edges between dissimilar nodes based on the cosine similarity between the connected nodes. As our proposed attack is based on injection, this defense is the most suitable for detecting the injection. If the features of the trigger are dissimilar and out of distribution for the original nodes in the clean graphs, edge pruning method can effectively detect and cut down the trigger part to make the graph clean benign again. For an edge between nodes (u,v), the cosine similarity can be formally computed by:

%\begin{equation}
%sim(u,v) = \frac{x_u.x_v}{||x_u|| ||x_v||}
%\end{equation}

%the amount of edges that will be pruned is controlled by similarity threshold parameter which will be set by the defender. The edges which have a lower similarity value than the threshold will be reduced.

\subsection{Experimental Setup}
To evaluate the robustness of our proposed multi-targeted attack \texttt{BadImplant} against defense mechanisms, we conduct comprehensive evaluation across all five datasets (MNIST, CIFAR-10, ENZYMES, Reddit-Multi-12k and Reddit-Multi-5k). For each dataset, we use the trained GCN model with three target classes and the default configuration described in Section \ref{subsec:exp_setting}. For randomized smoothing, we use 100 as the number of samples (n), meaning there will be 100 variants of the input graphs, and majority voting will provide the final prediction. We vary the magnitude of the gaussian noise by changing the values of noise standard deviation (\(\sigma\)) and observe the impact of the defense method on ASRs and CAs of the model. For fine pruning, we use 5-10\% of the dataset as the clean samples depending on the original dataset size. 

\subsection{Results and Discussion}
Figure \ref{fig:rs_defense_analysis}(a--e) illustrates the evaluation of our \texttt{BadImplant} against the randomized smoothing defense mechanism across five datasets with various noise intensity. The results exhibit the resilience of our proposed framework against noise-based defense, where the ASRs remain high despite noise is pumped to a level that substantially disrupts the model's primary classification objective. 

On MNIST dataset, the average ASR is above 90\% even when the noise severity degrades the clean accuracy from 90\% to 60\%. Similar pattern of ASR and CA was observed for CIFAR-10 and ENZYMES datasets. With the increased noise standard deviation, average ASR on CIFAR-10 remains around 80\%, despite the model's clean accuracy drops down to the half. As shown in Figure \ref{fig:RS_ENZYMES}, the ENZYMES dataset also exhibits comparable resilience, where the average ASR retains about 80\% yet clean accuracy declines to approximately 35\%. These results demonstrate that our \texttt{BadImplant} persists even when the model becomes malfunctioning with heavy injected noise. 

For those datasets consisting only topology (Reddit-Multi-12k and Reddit-Multi-5k), the node degree is utilized as the node feature. On both datasets, the ASRs remain remarkably unaffected with the introduction of the Gaussian noise. For instance, Figure \ref{fig:red_12k_rs} and \ref{fig:red_5k_rs} present the average ASR of around 99\% for both datasets across the entire noise standard deviation range (\(\sigma=\) 0 to 0.8). Clean accuracy does not drop with the increase of the magnitude of noise on Reddit-Multi-12k, but with less than 5\% drop on Reddit-Multi-5k. Considering that these two topology-based datasets relying on graph connectivity more than on node features, the resulting ASR and CA are more stable with increased noise injection. In summary, our \texttt{BadImplant} can effectively withstands noise-based defense mechanisms, achieving high ASRs despite the model's clean accuracy is significantly compromised.

To evaluate the robustness of our attack against the fine pruning-based defense, we systematically increase the pruning ratio to examine the impact on the ASRs and CAs of the backdoored GNN model. Figure \ref{fig:fp_defense_analysis}(a--e) illustrate the performance of our multi-targeted \texttt{BadImplant}. The backdoored model maintains high ASRs on all five datasets while models clean accuracies are hampered notably by the fine pruning. The average ASR on MNIST dataset exceeds 90\% with the clean accuracy dropped from 90\% to below 40\%. 

Consistent trends can be observed across the remaining datasets. The average ASR on CIFAR-10 remains above 80\% while the models' clean accuracy drops down to about 38\%. The average ASR on ENZYMES is above 80\% while the CA falls below 35\%. On the other hands, both Reddit-Multi-12k and Reddit-Multi-5k datasets tend to be more sensitive to fine pruning with sharply degrading CA as the pruning ratio increases. Nevertheless, the average ASRs are above 80\% on both Reddit datasets when the CAs are halved. Those experiments validate the robustness of our \texttt{BadImplant} against fine pruning across diverse graph datasets.

Overall, the experimental results demonstrate that both randomized smoothing and fine pruning fail to meaningfully disrupt our multi-targeted backdoor attack. Across all five datasets, the ASRs preserve significantly high in spite of degraded clean accuracy, suggesting that our \texttt{BadImplant} is robust and effective.

\section{Conclusion}\label{sec:con}

%We demonstrate that adversaries can execute highly effective sample-specific and multi-targeted backdoor attacks using white Gaussian noise triggers based on power spectral density. Our \texttt{NoiseAttack} achieves high attack success rates across three datasets and architectures in image classification while maintaining clean accuracy. \texttt{NoiseAttack} bypasses state-of-the-art detection and mitigation techniques, opening new research directions for both attacks and defenses.

In this work, we presented the first multi-targeted backdoor attack named \texttt{BadImplant} on graph classification. We introduce subgraph injection-based approach which achieves high attack success rate with minimal compromise on the clean accuracy. Our empirical experimental results demonstrate the superiority of our \texttt{BadImplant} over conventional subgraph replacement-based backdoor attack. The attack performance across five datasets and four model architectures exhibits the effectiveness and generalization capability of our attack. Results on maximum number of targets show that multiple triggers can successfully coexist without interfering with the prediction of other triggers. We further conduct extensive experiments to examine the effects of various attack design parameters on \texttt{BadImplant}, demonstrating the breadth of attack performance on various configurations. Moreover, \texttt{BadImplant} can preserve its effectiveness and resilience against two commonly certified defense methods. In conclusion, \texttt{BadImplant} provides a reliable and effective approach on backdooring graph neural networks. 

%\input{Sections/Limitations}

%\input{Sections/Ethics Statement}

%\bibliography{custom}

%{
%    \small
%    \bibliographystyle{ieeenat_fullname}
%    \bibliography{custom}
%}

\newpage

%% Loading bibliography style file
%\bibliographystyle{model1-num-names}
\bibliographystyle{plain}

% Loading bibliography database
\bibliography{bibtex/bib/IEEEexample}

\end{document}